\documentclass[fleqn,10pt]{wlscirep}
\usepackage[utf8]{inputenc}
\usepackage[T1]{fontenc}
\usepackage{lineno}
\usepackage{float}
\usepackage{amsmath}
\usepackage{eurosym}
\usepackage{svg}
\usepackage{float}
\usepackage{cleveref}

\title{Synthetic Image Rendering Solves Annotation Problem in Deep Learning Nanoparticle Segmentation}

\author[1, 2, *]{Leonid Mill}
\author[4]{David Wolff}
\author[3]{Nele Gerrits}
\author[5]{Patrick Philipp}
\author[4,7]{Lasse Kling}
\author[1, 4]{Florian Vollnhals}
\author[5]{Andrew Ignatenko}
\author[2]{Christian Jaremenko}
\author[2]{Yixing Huang}
\author[5]{Olivier De Castro}
\author[5]{Jean-Nicolas Audinot}
\author[3]{Inge Nelissen}
\author[5]{Tom Wirtz}
\author[2]{Andreas Maier}
\author[1, 4, 6, 7]{Silke Christiansen}
\affil[1]{Institute of Optics, Information and Photonics, Friedrich-Alexander-University Erlangen-Nuremberg, Erlangen, Germany}
\affil[2]{Pattern Recognition Lab, Friedrich-Alexander-University Erlangen-Nuremberg, Erlangen, Germany}
\affil[3]{Health Unit, Flemish Institute for Technological Research, Mol, Belgium}
\affil[4]{Institut für Nanotechnologie und korrelative Mikroskopie, Forchheim, Germany}
\affil[5]{Advanced Instrumentation for Ion Nano-Analytics, Materials Research and Technology Department, Luxembourg Institute of Science and Technology, Belvaux, Luxembourg}
\affil[6]{Physics Department, Free University, Berlin, Germany}
\affil[7]{Fraunhofer Institute for Ceramic Technologies and Systems, Dresden, Germany}
\affil[*]{leonid.mill@fau.de, schrist1@gwdg.de}


\begin{abstract}
Nanoparticles occur in various environments as a consequence of man-made processes, which raises concerns about their impact on the environment and human health. To allow for proper risk assessment, a precise and statistically relevant analysis of particle characteristics (such as e.g. size, shape and composition) is required that would greatly benefit from automated image analysis procedures. While deep learning shows impressive results in object detection tasks, its applicability is limited by the amount of representative, experimentally collected and manually annotated training data. Here, we present an elegant, flexible and versatile method to bypass this costly and tedious data acquisition process. We show that using a rendering software allows to generate realistic, synthetic training data to train a state-of-the art deep neural network. Using this approach, we derive a segmentation accuracy that is comparable to man-made annotations for toxicologically relevant metal-oxide nanoparticle ensembles which we chose as examples. Our study paves the way towards the use of deep learning for automated, high-throughput particle detection in a variety of imaging techniques such as microscopies and spectroscopies, for a wide variety of studies and applications, including the detection of plastic micro- and nanoparticles.
\end{abstract}
\begin{document}

\flushbottom
\maketitle

\section*{Introduction}
Nanoparticles are omnipresent in our daily lifes. They can be found in products ranging from cosmetics, textiles and foods, and are important in various technological fields such as energy, electronics, medicine and many more. \cite{vance2015nanotechnology, lohse2012applications, sun2007bright, dreaden2012golden, zhang2014near} However, as a consequence of industrial processes and man-made pollution, unwanted nanoparticle size distributions and concentrations \cite{bundschuh2018nanoparticles} give rise to concerns with respect to human health and environmental pollution. While the nanoparticles’ physicochemical properties (size, shape, surface chemistry, etc.) determine the quality of products \cite{kongkanand2008quantum, mackey2014most}, such characteristics are also important in order to evaluate the biological impact of nanoparticles  at a molecular, cellular and systemic level for any risk assessment for environmental and human health \cite{mulhopt2018characterization}. Characterizing nanoparticles in a dynamic context and on a case-by-case basis, microscopic imaging techniques including those that use focused electron or ion beams in scanning electron microscopes (SEMs) or Helium Ion Microscopes \cite{hlawacek2014helium} (HIMs) to generate nanometer scale spatial resolution are frequently applied in the scientific community. Given the substantial information content of digital images, these techniques often benefit from, or require, automated high-throughput data analysis that enables the accurate identification of large numbers of particles in a robust way.
\\
\\
Several approaches have been proposed for automated nanoparticle detection using traditional algorithms as well as machine learning techniques.\cite{Wang2016, Al-Dulaimi2017, ChiwooPark2012, Mirzaei2017, Meng2018} However, all of these state-of-the art approaches encounter major difficulties caused by irregular object patterns and noise \cite{Riccio2019}, or they rely on hand-crafted features for particle shapes \cite{Meng2018, Mirzaei2017}, which impair the generalization potential of such algorithms for the characterization of arbitrary nanoparticles or heterogeneous nanoparticle ensembles. In order to handle various sizes, shapes and distributions of nanoparticles more sophisticated image analysis approaches are required.
\\
\\
With the recent advancements in machine learning and mainly deep learning \cite{Lecun2015}, deep convolutional neural networks (CNNs) \cite{He2017, simonyan2014very, he2016deep, Redmon2016, ronneberger2015u, Falk2019} have been developed, which are able to learn from data sets containing millions of images \cite{JiaDeng2009} to resolve object detection tasks. When trained on such big data sets, CNNs are able to achieve task-relevant object detection performances that are comparable or even superior to the capabilities of humans \cite{Silver2016, krizhevsky2012imagenet}. While artificial neural networks have already been used for the classification and inverse design of nanoparticles \cite{malkiel2018plasmonic, peurifoy2018nanophotonic}, the key problem to use deep learning for nanoparticle detection is, in general, the large amount of data needed to train such networks. The main difficulty lies in the acquisition of a representative data set of nanoparticle images which ideally contain various sizes, shapes and distributions for a variety of nanoparticle types. Additionally, manual annotation of the acquired data is mandatory to obtain the so called `ground truth' or `labels', which is, in general, error-prone, time-consuming and consequently costly. Although approaches exist such as `precision learning' \cite{Maier2019} or `transfer learning' \cite{pan2009survey} that reliably work with a substantially reduced amount of training data, a certain data set size is still required that incorporates human effort, i.e. for manual annotation of the acquired data.
\\
\\
To overcome the aforementioned data limitations, we have developed a semi-automated data synthesis scheme using an open-source rendering software. Particularly, we use Blender \cite{blender2018} in this work, in order to generate synthetic data based on very limited amounts of real microscopic data sets. The proposed workflow enables the generation of a virtually unlimited number of synthetic and photo-realistic microscopic nanoparticle images comprising various types of particles with different sizes, shapes, compositions and three-dimensional distributions. Thereby, the respective synthetic ground truth segmentation mask is automatically derived for each generated image. We show that the synthetic data is sufficient in terms of realism and size to successfully train a deep learning model for segmentation which is also able to operate on real data. As a demonstration example, we generate photo-realistic synthetic data sets of HIM images of metal-oxid and metal nanoparticles (SiO$_{2}$, TiO$_{2}$ and Ag) that serve as training data for a state-of-the art \cite{hemelings2019artery, Falk2019} deep CNN so called U-Net \cite{ronneberger2015u}. For experimental validation, manual annotations of the HIM images have been carried out by experts. We show that training the CNN on synthetic data yields similar segmentation accuracy as using
real HIM images. Furthermore, we demonstrate the applicability of the proposed method for images containing very complex shaped and distributed Ag particles, nanorods and nanowires.

\section*{Novel, semi-automated nanoparticle segmentation workflow}
\begin{figure}[!htb]
\centering
\includegraphics[scale=0.75]{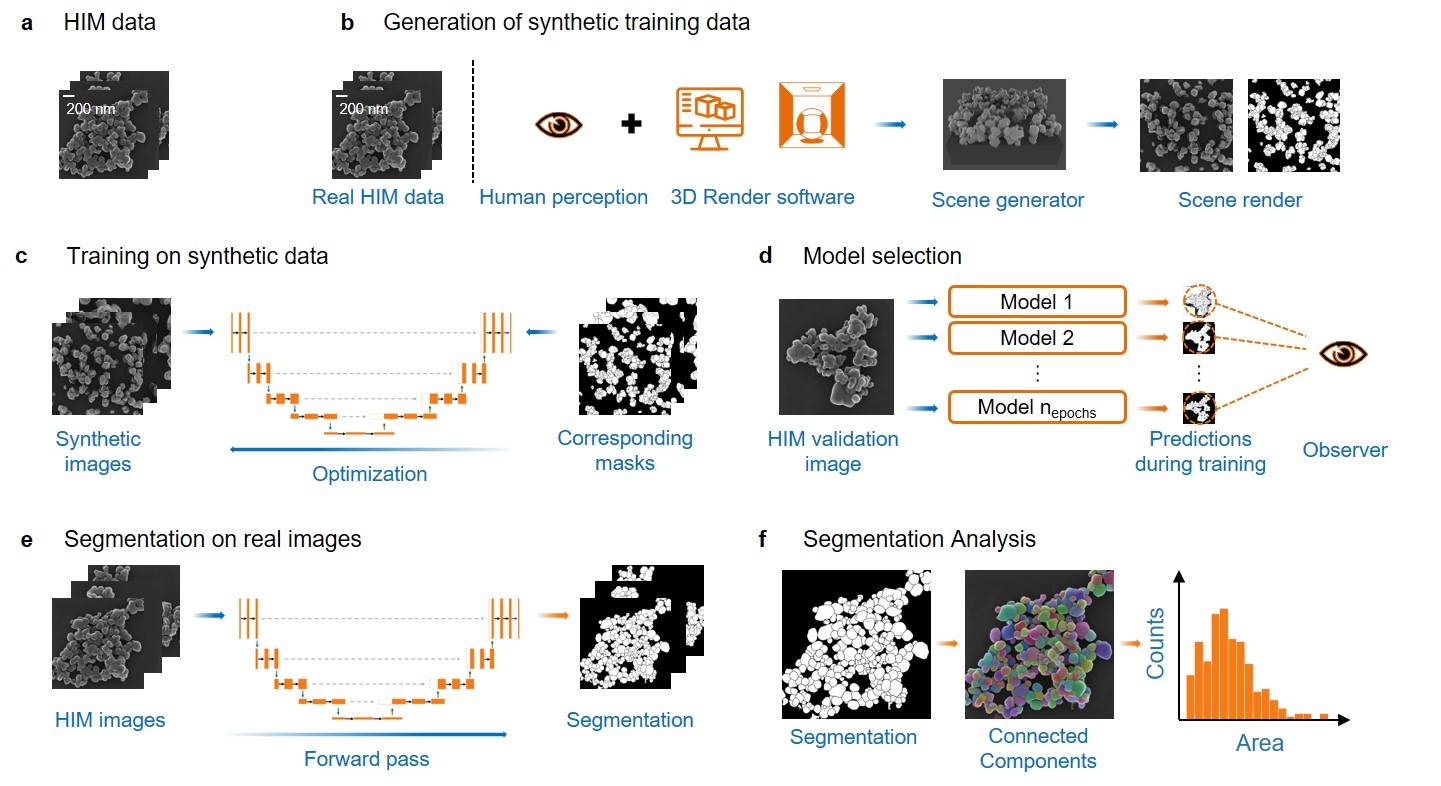}
\caption{\textbf{Illustration of the nanoparticle segmentation workflow for real Helium ion microscope (HIM) nanoparticle images using a deep convolutional neural network (CNN) and synthetic training data. a}, HIM images of TiO$_{2}$ nanoparticle ensembles on a silicon (100) substrate surface. \textbf{b}, A semi-automated procedure to synthesize HIM images of TiO$_{2}$ particles based on the knowledge of a limited number of real HIM reference images. Using a render software (here Blender \cite{blender2018}) in combination with the information on particle size, shape and distribution gained from the reference images it is possible to create synthetic images that mimic real HIM data realistically. During this process, the respective ground truth labels for the synthetic TiO$_{2}$ image data are also generated in a fully automated manner. \textbf{c}, The synthetic data set is used to train a deep CNN (here the U-Net \cite{ronneberger2015u}) for automated particle segmentation and subsequent quantitative, statistical assessment. \textbf{d}, After training the CNN using synthetic data, the model which predicts the most accurate segmentation on a real validation image is selected for further processing. \textbf{e}, The best CNN model is used to predict the segmentation masks for all experimental microscopy images. \textbf{f}, The accurate prediction of the model permits further statistical image analysis.}
\label{fig:workflow}
\end{figure}

Our goal is to extract statistical information on nanoparticle morphology with respect to size, shape and distribution from high resolution microscopy images acquired with electron- or ion microscopes. However, to guide particle segmentation by deep CNNs, a representative amount of training data is required. This is in general a true limitation due to the fact that microscopy image data is usually not available in sufficiently large amounts to be suitable for robust CNN training purposes. Moreover, we want to overcome the manual image annotation process which is very time-consuming and error-prone. Here, we propose a novel workflow that overcomes the data problem for the deep learning based analysis of particle images. The workflow relies on a very limited number of real reference images, e.g. out of electron or ion microscopes, which serve as a blue print for the semi-automated photo-realistic synthetic data generation using a rendering software e.g. such as Blender. The proposed nanoparticle segmentation workflow is composed of the steps shown in Fig. \ref{fig:workflow}. Images of TiO$_{2}$ nanoparticles on a silicon wafer surface, taken by a Helium ion microscope (HIM) serve as examples for the demonstration of the procedure and the quality of the statistical evaluation of the nanoparticle properties.
\\
\\
The starting point of the segmentation workflow is the acquisition of a limited number of high resolution HIM images of TiO$_{2}$ nanoparticles, spread on a (100) silicon wafer. Fig.~\ref{fig:workflow}a shows an example of such an HIM image. Based on a number of images of that type (see Supplementary Fig. \ref{supp_fig:tio2_raw}) a human is able to obtain an  overall impression of the average size, shape, potential faceting and orientation of the TiO$_{2}$ nanoparticles with respect to one another as well as the degree of homogeneity of the particle distribution within those images (Fig.~\ref{fig:workflow}b). The knowledge gained from the initial manual assessment of the nanoparticle data is used to reproduce such features virtually using a rendering software, e.g. Blender \cite{blender2018}, and to automatically generate 3D scenes filled with synthetic TiO$_{2}$ particle ensembles. During this automated rendering process, details of the individual particles as well as the overall 2D or 3D arrangements on the substrate surface, including potential image artefacts (i.e. such as dirt which may arise from sample preparation procedures), are included in virtual Blender scenes to mimic real TiO$_{2}$ particle ensembles. An automatically generated TiO$_2$ scene is demonstrated in Supplementary Fig. \ref{supp_fig:tio2_template_particles}. Details on the automatic Blender scene generation process are provided in the Methods section. Subsequently, after an user-defined number of scenes is generated, each Blender scene is automatically being rendered twice. The first rendering process computes a photo-realistic synthetic microscope image of the TiO$_{2}$ nanoparticles and potential artefacts, while a subsequent render produces the respective error-free ground truth label image (Fig.~\ref{fig:workflow}b).
\\
\\
With a sufficient amount of realistic synthetic data, a deep CNN can be properly trained on the nanoparticle segmentation task. In our workflow, we use the so called U-Net \cite{ronneberger2015u}, a the state-of-the art \cite{hemelings2019artery, Falk2019} deep CNN, which was originally proposed and successfully applied for the segmentation of cells in transmission electron microscope images. Fig.~\ref{fig:workflow}c shows the U-Net comprising an U-shaped encoder-decoder architecture. The orange boxes represent multi-dimensional feature maps extracted by different convolution layers, while the dashed gray arrows correspond to skip-connections. A detailed description of the architecture is provided by Ronneberger et al. \cite{ronneberger2015u}. During the U-Net training process, the network iteratively optimizes its internal parameters to learn the segmentation of nanoparticles based on our synthetic microscopic images by predicting corresponding segmentation masks. To prevent the network from overfitting on the synthetic data, at least one real HIM 'validation' image is used during training. In machine learning, 'validation data' is used during the training process to evaluate the model performance on hold-out data which is not part of the training data set. Therefore, after each complete training iteration, also referred to as an 'epoch' in machine learning, the model predicts a segmentation for the validation image (Fig. \ref{fig:workflow}d). After the CNN training process, a human observer visually evaluates all predicted segmentation masks for the validation image and selects the model which provides the qualitatively most accurate particle segmentation. Subsequently, the selected model is used to segment all remaining real microscopic (here from a HIM) images (Fig.~\ref{fig:workflow}e), while subsequently, all sorts of statistical and quantitative information can be deduced for the microscopic HIM images of interest (Fig. \ref{fig:workflow}f).

\section*{Synthetic image quality}
We used the aforementioned semi-automated data generation scheme to generate synthetic microscope images such as the HIM images of SiO$_{2}$, TiO$_{2}$ and silver nanoparticles of varying overall morphology. 
\begin{figure}[ht!]
\centering
\includegraphics[scale=0.82]{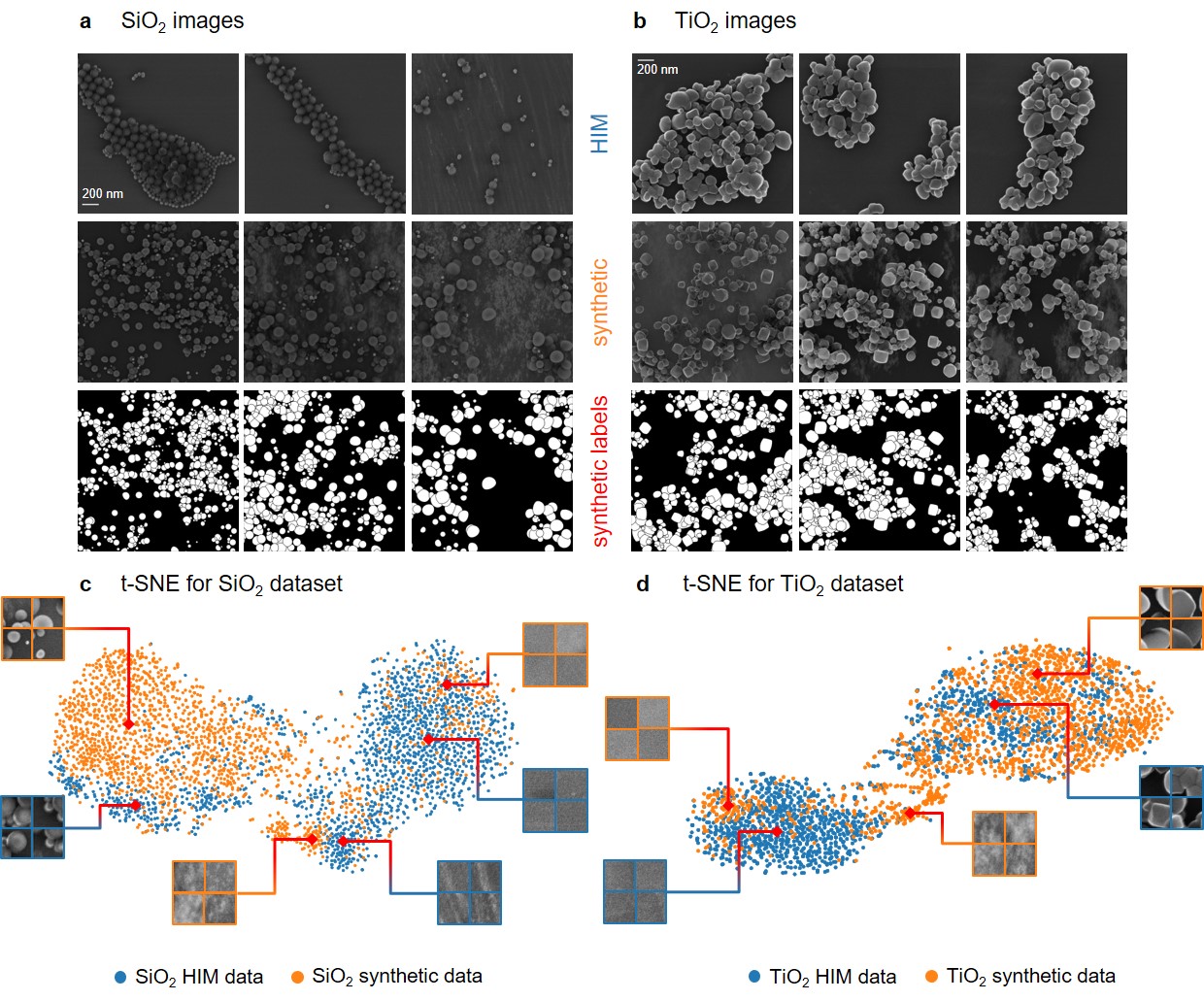}
\caption{\textbf{a-b: Comparison of metal-oxide nanoparticles on silicon wafers imaged by a HIM (top row) with synthetically generated images using the proposed semi-automated rendering process (middle row) with its respective synthetic labels (bottom row).} \textbf{a}, Spherical SiO$_{2}$ nanoparticles with various dimensions on a silicon wafer surface arranged in clusters or a certain degree of alignment in close vicinity to neighboring particles. Also, note the dirt smear on the silicon wafer surface in the rightmost image, which was also covered in the simulation. The synthetic images in the middle row on the contrary show a more statistical distribution of particles compared to the real images, which also include the surface impurity feature. \textbf{b}, Microscopic HIM images of complex shaped and distributed TiO$_{2}$ nanoparticles. \textbf{c-d: t-SNE visualization of the SiO$_{2}$ and TiO$_{2}$ data sets.} Each data point corresponds to an image patch of identical size (144$\times$144 px) of real (blue) and synthetic (orange) images. Both t-SNE plots show a distinct separation between background and foreground (particle) data points, while nearby and overlapping
points indicate the high similarity between synthetic and real data. \textbf{c}, The t-SNE result for the SiO$_{2}$ data set shows two clusters of particle images (left cluster), background images (right cluster) with an additional cluster for image patches, that show surface impurity. The small overlap between real and synthetic SiO$_{2}$ particle images indicates that the synthetic images are similar, but
do not represent a perfect match of the target HIM data \textbf{d}, t-SNE plot for the TiO2 particles. The large overlap between synthetic and real particle images (right cluster) indicates that the synthetic images represent the real TiO$_{2}$ HIM data quite well.}
\label{fig:real_vs_syn_t_SNE}
\end{figure}
A comprehensive set of real HIM images of such nanoparticles are shown in the Supplementary Section in figures Figs.~\ref{supp_fig:sio2_raw}-\ref{supp_fig:ag_raw}. However, since no human derived manual annotation was performed for silver nanoparticle ensembles in HIM images due to the complexity of the data, in the following we will set our focus on SiO$_{2}$ and TiO$_{2}$ nanoparticles for the quantitative and qualitative evaluation. Fig. \ref{fig:real_vs_syn_t_SNE} provides a direct comparison of SiO$_{2}$ (Fig. \ref{fig:real_vs_syn_t_SNE}a) and TiO$_{2}$ (Fig. \ref{fig:real_vs_syn_t_SNE}b) particles in HIM images (top row) with the corresponding synthetic photo-realistic images (middle row) and its respective synthetic labels (bottom row). 
\\
\\
The real SiO$_{2}$ particles essentially show a very regular, spherical shape, which was also considered in the corresponding synthetic SiO$_{2}$ simulations (see Supplementary Fig. \ref{supp_fig:sio2_template_particles}). The arrangement of the SiO$_{2}$ particles in the real HIM images is not random, the particles align in close vicinity, ideally in clusters or rows of particles. In contrast, the particles in the synthetic images are distributed randomly (see Supplementary Fig. \ref{supp_fig:sio2_render_scene}) throughout the substrate in order to generate very complex distribution in contrast to the real images and thus increase the variance in the data. The TiO$_{2}$ particles on the other hand form agglomerates, are arranged in sheets and have complex shapes. Therefore, in order to cover these characteristics in the synthetic data, we designed a range (we chose four) of virtual 3d so called 'template objects' (see Supplementary Fig. \ref{supp_fig:tio2_template_particles}) that were randomly combined to complex agglomerates in the TiO$_{2}$ render scenes during the automated image creation process (see Supplementary Fig. \ref{supp_fig:tio2_render_scene}). For both particle types (SiO$_{2}$ and TiO$_{2}$), we were able to mimic the characteristics of all relevant particle characteristics in terms of shape and distribution while comprising a more complex localization of particles. Additionally, due to the presence of substrate surface impurity ('dirt') that stems from imperfect sample preparation, we extended our simulation with a randomized impurity texture creation (see Fig.~\ref{fig:workflow}b and Supplementary Figs.~\ref{supp_fig:sio2_render_distibution_dirt},\ref{supp_fig:tio2_render_distibution_dirt}) to account for such features. Further details on the impurity texture are provided in the Methods section. Further details on the impurity texture are provided in the Methods section. Note that in contrast to manually annotated images, our synthetic labels do not contain any mislabeling and provide error-free, consistent particle contour lines as displayed in Fig.~\ref{fig:real_vs_syn_t_SNE}a-b (bottom row). Further synthetic images of SiO$_{2}$ and TiO$_{2}$, as well as the synthetic representation of more complex silver (Ag) nanoparticles that align in from of particle wires, are shown in Supplementary Figs. \ref{supp_fig:sio2_synthetic}-\ref{supp_fig:ag_synthetic}.
\\
\\
To assess the similarity between the real and synthetic HIM nanoparticle images, we analyzed the SiO$_{2}$ and TiO$_{2}$ data sets using the t-distributed stochastic neighbor embedding \cite{maaten2008tsne} (t-SNE). t-SNE is an unsupervised dimensionality reduction method which is primarily used to visualize high-dimensional data. In simpler terms, t-SNE provides for an idea of how data is arranged in a high-dimensional space. In general, nearby and overlapping data points in a t-SNE plot indicate similar data, while distant data points correspond to significant differences in the data. The resulting t-SNE plots for the SiO$_{2}$ and TiO$_{2}$ nanoparticles are depicted in Fig.~\ref{fig:real_vs_syn_t_SNE}c and d, respectively. Each scatter point corresponds to an image patch or sample of the size of 144$\times$144 px (with one pixel covering an area of 0.976$\times$0.976 nm$^2$) of a real (blue) or synthetic (orange) HIM image. More technical details on the sample extraction and data visualization methods are provided in the Methods section. In both t-SNE plots, a distinct separation between particle and background images can be observed. The nearby and overlapping blue and orange points indicate the overall high similarity between synthetic and real images. Moreover, for the SiO$_{2}$ data, the samples that contain substrate surface impurity in both real and synthetic data points show a high comparability which suggest a very accurate dirt simulation. 
However, the small overlap between real and synthetic SiO$_{2}$ particle images indicates that although the synthetic images are comparable to the real HIM images, the simulation does not perfectly match the target HIM data. While the real data shows a higher variance of background images in terms of noise distribution and pixel intensities, the opposite holds for the the synthetic particle data, which contains presumably more complex particle shapes, distributions and variable contrasts. Additionally, in both plots an imbalance between synthetic and real particle and background data points can be observed. In this context, synthetic SiO$_{2}$ and TiO$_{2}$ images contain significantly more particle than background data. This is an intended and expected behaviour which results from our assumption that the simulation of random particle distributions with a high number of particles leads to a high variance of complex synthetic particle ensembles. As a consequence, it is expected that a deep CNN trained on complex synthetic data is more robust to variations in real images.
\begin{figure}
\centering
\includegraphics[scale=0.8]{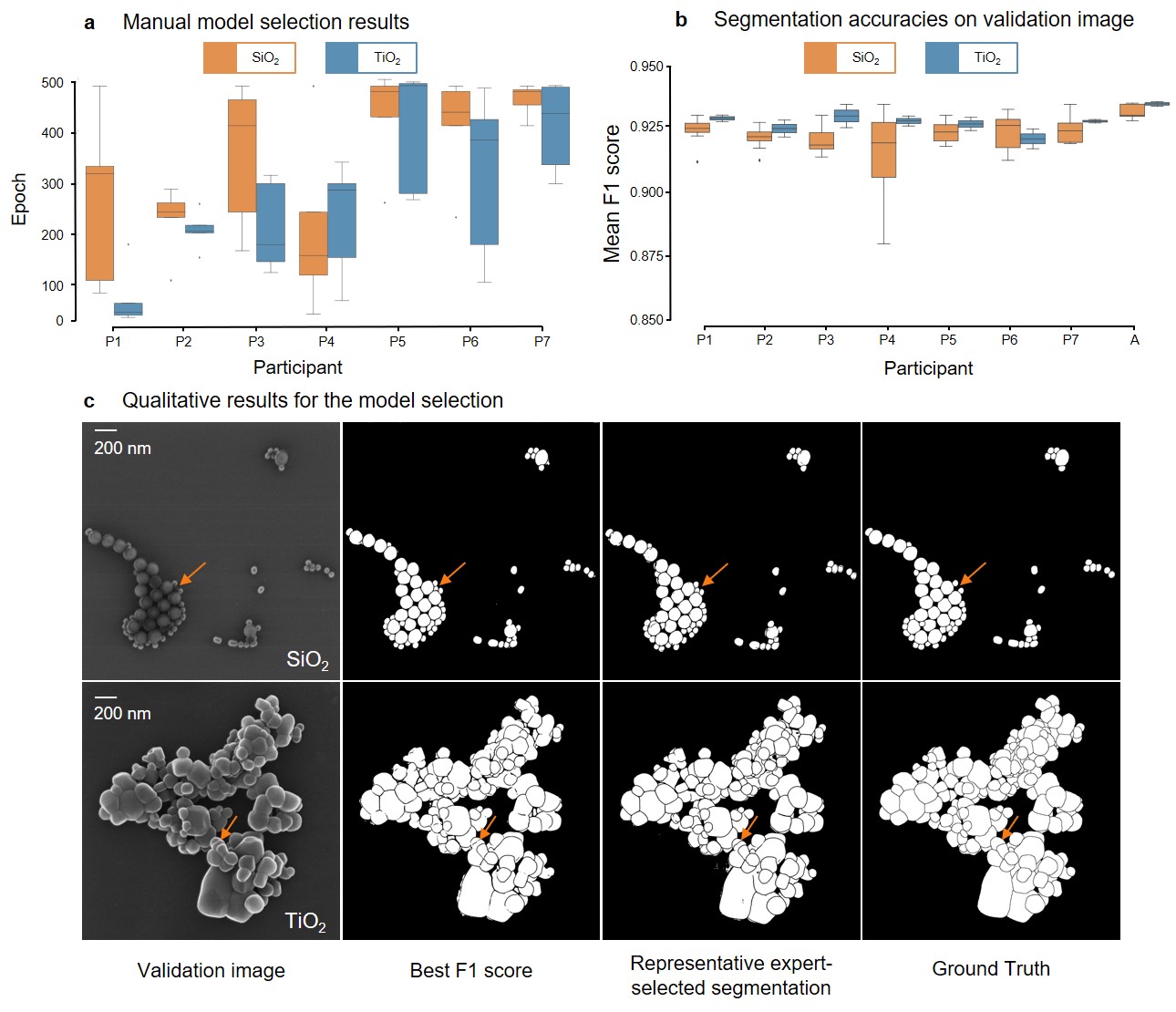}
\caption{\textbf{Comparison of two model selection approaches to find the best U-Net model based on real HIM validation images of SiO$_{2}$ and TiO$_{2}$ nanoparticles.} While the `analytical' approach relies on a manual annotation of the validation image in order to find the best model, the second approach is based on the qualitative (visual) evaluation of the segmentation masks predicted by each model. To compare both methods, a user study with seven participants was conducted with non-experts (P1-P4) and experts (P5-6) for segmentation. According to its visual perception, each participant had to choose five best network models based on the model's segmentation performance for a real HIM validation image. \textbf{a}, Visualization of the five models, defined by the epoch number, that were selected by each of the study participants. \textbf{b}, F1 scores (y-axis) averaged over the 5 chosen models by each participant (P1-P7), using the manually annotated GT as reference segmentation for the validation image. Additionally, the diagram shows the mean F1 score for the `analytical' model selection approach, for comparison reasons also averaged over the 5 best models. \textbf{c}, Visualization of the validation images SiO$_{2}$ (top row) and TiO$_{2}$ (bottom row), the corresponding segmentation masks predicted by the model with the best F1 score selected by the `analytical' approach, the segmentation of a representative expert-selected model as well as the manual GT reference segmentation. Although the analytically selected model and the representative expert-selected model both provide an accurate overall segmentation, the expert-selected model is more suitable to separate individual particles (TiO$_{2}$) which is mandatory for statistical particle analysis. The orange arrows emphasize segmented particles by the U-Net that do not occur in the manually annotated ground truth image.}
\label{fig:study_evaluation_validation}
\end{figure}

\section*{Model selection}
Since the U-Net was trained on synthetic data only (Fig.~\ref{fig:workflow}c) real validation data is needed to assess the model performance for real HIM particle data. To select the best model, we compared two approaches for the model selection in a user study with 7 participants, consisting of three experts on segmentation and four non-experts. The first approach, referred to as 'analytical', requires the time-consuming manual annotation of at least one real HIM validation image, which is subsequently used as ground truth (GT) to select the best model based on the highest F1 score \cite{dice1945measures}. The F1 score is a metric which is often used to assess the segmentation accuracy of a model compared to a reference segmentation. The second approach relies on a human observer to choose the best model according to the qualitatively (visually) best segmentation performance (Fig.~\ref{fig:workflow}d). Further details on the user study and model selection are presented in the Methods section. Fig. \ref{fig:study_evaluation_validation}a illustrates the models, defined by the epoch number, that were selected by each participant for the SiO$_{2}$ (orange) and TiO$_{2}$ (blue) data sets, respectively.
The plot illustrates that the expert group (P5-P7) mostly selected models towards the end of the training process, while no significant trend could be observed for the non-expert group (P1-P4). A detailed qualitative evaluation of the selected segmentation masks showed that in contrast to the non-expert group, the experts tend to choose models that demonstrated a higher capability of separating particles in more detail, producing more accurate contour lines, whereas non-experts relied on the overall segmentation performance only. Fig.~\ref{fig:study_evaluation_validation}b visualizes the segmentation accuracies of the selected (manually and 'analytically') five best models on the validation image based on the mean F1 score.
\\
\\
For both particle types (SiO$_{2}$ and TiO$_{2}$), the non-expert group (P1-P4) as well as the experts (P5-P7) consistently selected models that provided almost perfect segmentation results on the validation image (SiO$_{2}$ - F1 mean (SD): 0.92 (0.01); TiO$_{2}$ - F1~mean~(SD):~0.93~(0.004)). However, the models selected by the 'analytical' approach demonstrate marginally more accurate segmentation performances (SiO$_{2}$ - F1 mean (SD): 0.93 (0.001); TiO$_{2}$ - F1 mean (SD): 0.93 (0.001)). A qualitative (visual) comparison between the two approaches is displayed in Fig.~\ref{fig:study_evaluation_validation}c. This figure visualizes the validation images, the segmentation masks predicted by the model with the best F1 score according to the 'analytical' approach, the segmentation of a representative expert-selected model as well as the manual GT reference segmentation. Both approaches resulted in a model selection that showed a very high overall segmentation accuracy on the validation image compared to the GT reference. However, the model selected by the 'analytical' approach lacks a distinct separation of individual TiO$_{2}$ particles, while the expert implicitly selected a model that accounted for this behaviour. Moreover, the expert-selected model produces more precise particle contour lines over the 'analytical' approach. Since statistical analysis of nanoparticle images requires an accurate separation of particles, we further rely on expert-selected models for the particle analysis in the next section. Moreover, we want to emphasize that a human-based model selection is preferable due to the time-consuming and error-prone process of manually annotating a validation image which is required for the 'analytical' approach. Additionally, it was observed that training on synthetic images also resulted in a robust particle segmentation for the validation images, even for non-trivial cases as illustrated by orange arrows in Fig. \ref{fig:study_evaluation_validation}c, which highlight particles that are segmented as individual particles by the model but are missing in GT segmentation.

\section*{Segmentation performance and statistical particle analysis}
Finally, after the model selection, the best model is used for the segmentation of all remaining real HIM images (Fig~\ref{fig:workflow}e). In order to compare the U-Net that was trained on simulated data (U-Net\textsubscript{sim}) with a baseline performance, we manually annotated all real SiO$_{2}$ and TiO$_{2}$ HIM images and trained a U-Net on real data (U-Net\textsubscript{real}) for each particle type, respectively. Note that the manual annotation of SiO$_{2}$ and TiO$_{2}$ HIM images was performed by one individual each. Due to the small number of real HIM images (SiO$_{2}$: 9, TiO$_{2}$: 8) in comparison to the number of synthetically generated images (SiO$_{2}$: 180, TiO$_{2}$: 180), U-Net\textsubscript{real} was trained in a leave-one-out cross-validation \cite{BROWNE2000108} setup to assess its general segmentation performance. Further details on the training and the post-processing used to enhance the segmentation quality are provided in the Methods section.
\begin{figure}[t!]
\centering
\includegraphics[scale=1]{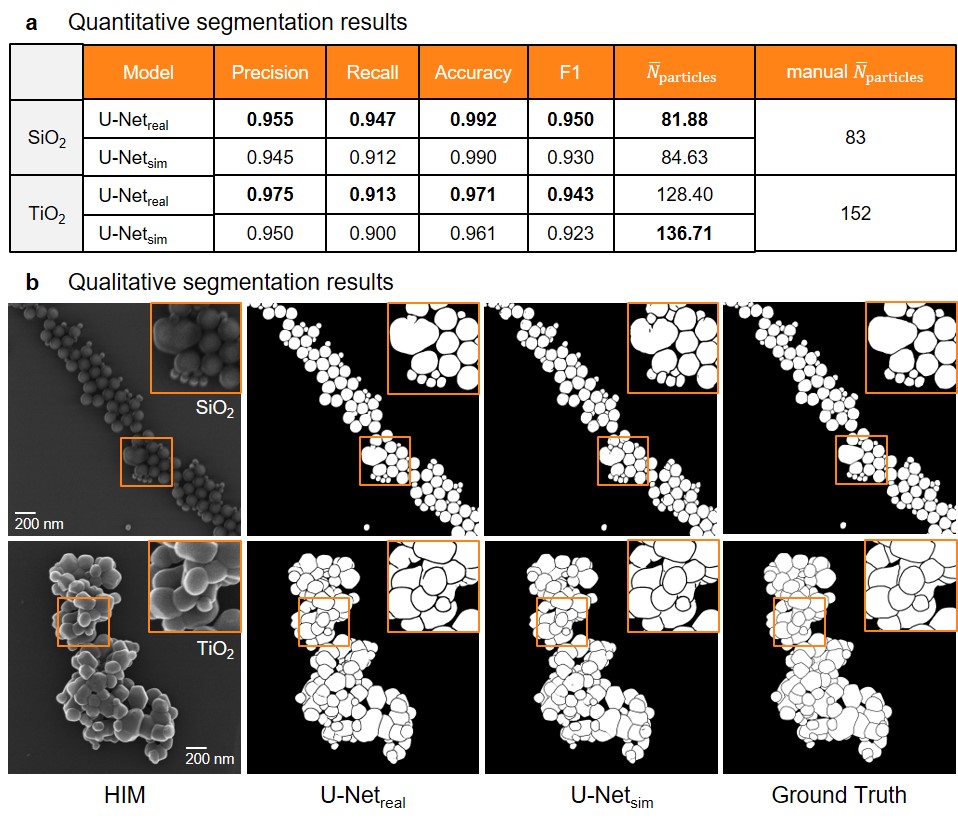}
\caption{\textbf{Segmentation results for the U-Net trained on real data (U-Net\textsubscript{real}) compared with the performance of the U-Net which was trained on simulated data (U-Net\textsubscript{sim}). a}, Quantitative segmentation results using precision, recall, accuracy and the F1 score as evaluation metrics (see Methods for details). The average number of particles detected by the models is denoted as '$\overline{N}\textsubscript{particles}$', while the number of individual particles that are present in the manual GT annotation is given by 'manual $\overline{N}\textsubscript{particles}$'.
\textbf{b}, Qualitative segmentation results for an exemplary selected real SiO$_{2}$ (top row) and TiO$_{2}$ (bottom row) HIM image. Both, the quantitative as well as qualitative results show almost perfect segmentations predicted by U-Net\textsubscript{real} and U-Net\textsubscript{sim} for both particle types compared to the manually annotated ground truth segmentation indicating the human-comparable segmentation capabilities of the models.}
\label{fig:segmentation_evaluation}
\end{figure}
\\
\begin{figure}[t!]
\centering
\includegraphics[scale=0.86]{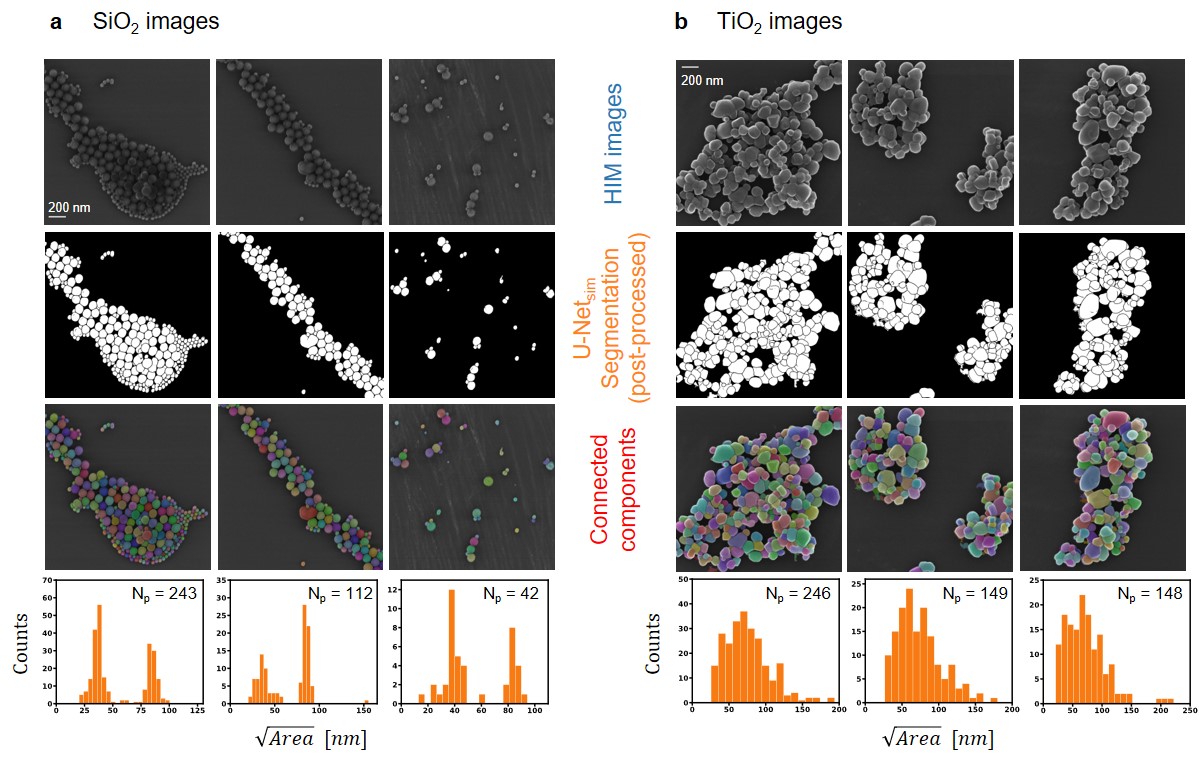}
\caption{\textbf{Analysis of SiO$_{2}$ (a) and TiO$_{2}$ (b) nanoparticle images.} The top row shows raw HIM image data, whereas the second row visualizes the post-processed segmentation prediction of the U-Net\textsubscript{sim}, which was trained on synthetic data only. The third row displays an overlay of the original HIM image with a connected component labeling (CCL) of the post-processed segmentation. In a CCL each color is associated with an individual particle. Note that due to the limited number of colors used in a CCL, neighboring particles may occur in the same color although being separated, individual particles. The bottom row provides a particle size distribution, based on the accurate U-Net\textsubscript{sim} prediction, for each HIM image which highlights the particle frequency with respect to the square-root of the particle area size denoted nanometer (nm). N\textsubscript{p} represents the number of particles. While the SiO$_{2}$ histograms reveal a bimodal character, separating the two groups of small and large particles, TiO$_{2}$ histograms are characterized by a log-normal distribution.}
\label{fig:segmentation_analysis}
\end{figure}
\\
Fig. \ref{fig:segmentation_evaluation}a provides a quantitative comparison of U-Net\textsubscript{real} and U-Net\textsubscript{sim} for the SiO$_{2}$ and TiO$_{2}$ HIM data. According to precision, recall, accuracy and F1 score, U-Net\textsubscript{real} (F1 score SiO$_{2}$: 0.950, F1 score TiO$_{2}$: 0.943) marginally outperforms the U-Net\textsubscript{sim} (F1 score SiO$_{2}$: 0.930, F1 score TiO$_{2}$: 0.923) for both particle types. However, both models provide nearly perfect segmentations. Additionally, for the SiO$_{2}$ data, the average number of individual particles ($\overline{N}\textsubscript{particles}$) detected by U-Net\textsubscript{real} (81.88) and U-Net\textsubscript{sim} (84.63) matches almost ideally the average number of 83 particles (manual $\overline{N}\textsubscript{particles}$) that are present in the manually annotated SiO$_{2}$ GT images. On the other hand, while the number of TiO$_{2}$ particles identified by U-Net\textsubscript{sim} (136.71) is more accurate than the number of particles detected by U-Net\textsubscript{real} (128.40), a considerable difference to manual $\overline{N}\textsubscript{particles}$ (152) can be observed. An explanation for this finding lies in the post-processing used to improve the segmentation output of the U-Nets (U-Net\textsubscript{real} as well as U-Net\textsubscript{sim}). It removes small objects below a certain (pixel) area size using a morphological operation called 'area opening', which results in a smaller amount of particles that are present in the segmentation masks after the post-processing (see Supplementary Figs.~\ref{supp_fig:tio2_histograms_1}-\ref{supp_fig:tio2_histograms_4}). Fig. \ref{fig:segmentation_evaluation}b visualizes the qualitative segmentation results for an exemplary selected real SiO$_{2}$ (top row) and TiO$_{2}$ (bottom row) HIM image. According to the visual impression, U-Net\textsubscript{real} and U-Net\textsubscript{sim} provide almost perfect segmentations in comparison to the manually annotated GT segmentation, which also reflect the quantitative results. A direct comparison of the qualitative segmentation results for all real HIM images is illustrated in the Supplementary  Figs.~\ref{supp_fig:sio2_histograms_1}-\ref{supp_fig:tio2_histograms_4}, including particle size distributions for the GT reference as well as for the segmentations predicted by U-Net\textsubscript{real} and U-Net\textsubscript{sim}. Both, the quantitative as well as qualitative results indicate the human-comparable segmentation capability of U-Net\textsubscript{real} and U-Net\textsubscript{sim}. At this point, we want to highlight the extraordinary results produced by U-Net\textsubscript{sim}. Although trained on synthetic data only, it is able to achieve impressive segmentation accuracies on real HIM data, which is not only comparable to a U-Net trained on real data but also to the segmentation capability of humans.
\\
\\
Due to the human-comparable segmentation accuracy of the U-Net\textsubscript{sim}, it is possible to derive various features of real-world particles including their sizes, shapes, localization, or distributions using a connected component analysis (CCA). Therefore, we analysed the segmentations of U-Net\textsubscript{sim} using a CCA to assess the particle size distributions of the real HIM SiO$_{2}$ and TiO$_{2}$ images as depicted in Fig.~\ref{fig:segmentation_analysis}a-b. A direct comparison of the particle size distributions for all HIM images based on the segmentations of U-Net\textsubscript{real}, U-Net\textsubscript{sim} and the manual annotations is visualized in the Supplementary Figs.~\ref{supp_fig:sio2_histograms_1}-\ref{supp_fig:tio2_histograms_4}. While the top row shows raw HIM SiO$_{2}$ and TiO$_{2}$ images, the second row displays the corresponding post-processed U-Net\textsubscript{sim} segmentations. The third row visualizes an overlay of the original HIM images with a connected component labeling (CCL) based on the U-Net\textsubscript{sim} segmentation. In a CCL each color is associated with an individual particle. Note that due to the limited number of colors used in a CCL, neighboring particles may occur in the same color although being separated, individual particles. The CCL results for all HIM images, including the complex Ag data, are provided in Supplementary Figs.~\ref{supp_fig:sio2_seg_1}-\ref{supp_fig:ag_seg_3}. Using the CCA, we can derive accurate particle size distributions (Fig. \ref{fig:segmentation_analysis}, bottom row), which highlight the number of particles with respect to the square-root of the particle area size in nanometer (nm). While the SiO$_{2}$ histograms (Fig. \ref{fig:segmentation_analysis}a) reveal a bimodal character, separating the two groups of small and large particles, TiO$_{2}$ histograms (Fig. \ref{fig:segmentation_analysis}b) are characterized by a log-normal distribution with a modal value of approximately 60 nm. Moreover, due to the accurately simulated substrate surface impurity in the synthetic training data, image artefacts as a result of sample preparation related dirt in real SiO$_{2}$ HIM images does not affect the segmentation accuracy of the U-Net\textsubscript{sim} which has properly been trained to deal with this type of image features.

\section*{Conclusions}
In summary, we propose a segmentation workflow for complex nanoparticles in high resolution microscopic images which relies on deep learning in combination with a semi-automated synthetic data generation pipeline based on photo-realistic rendering. With this approach an unlimited number of realistic synthetic images can be created with its respective error-free ground truth labels which subsequently can be used to effectively train a deep CNN. The so trained CNN can subsequently be used to accurately segment nanoparticles in microscope images. We have demonstrated the applicability of the workflow on experimental HIM data sets of SiO$_{2}$ and TiO$_{2}$ nanoparticles spread on a silicon (100) wafer surface. We showed that the segmentation accuracy of a state-of-the art deep CNN trained only on synthetic data was comparable to segmentations carried out by microscopy experts. Moreover, we have demonstrated quantitative and statistical analysis of  microscopy imaged nanoparticles based on the automated deep learning CNN segmentation. 
\\
\\
We are confident that the method presented in this work has the potential to solve the training data bottleneck and the annotation problem for automated image analysis approaches and paves the way towards a wider use of deep learning in a variety of microscopy applications. It permits the implementation of automated high-throughput particle segmentation and characterization methods for all sorts of applications based on microscopic images. This can be particularly relevant for studies related to nano-toxicology and other fields such as nano- and bio-medicine, consumer product efficacy testing, and anti-counterfeiting. The time-consuming, costly and error-prone process of acquiring and manually annotating a representative amount of real data in order to use the power of deep learning is overcome by the proposed approach. Yet, the semi-automated data generation procedure still relies on human input (i.e. for the design of all relevant 3D scene parameters, template particles, shaders etc.) in order to achieve photo-realistic renders. However, with the recent achievements in differentiable rendering \cite{nimier2019mitsuba, loubet2019reparameterizing}, we assume that it will be possible to further automate this process.
\bibliography{references}
\clearpage
\newpage
\section*{Methods}

\subsection*{Sample preparation and data acquisition}
SiO$_{2}$ nanoparticles with two different diameters and food grade TiO$_{2}$ nanoparticles (E171) with a size distribution of 20 to 240 nm, both deposited on silicon chips (reference AGAR: G3390-10), have been obtained from the “Laboratoire National de métrologie et d’Essais”. The Polyvinylpyrrolidon (PVP)-coated Ag nanowires (PL-AgW50-10mg) with an average diameter of 40-50~nm and a length of up to 50 $\mu$m have been purchased from PlasmaChem GmbH (Berlin, Germany). A stable suspension was obtained by using a standard protocol \cite{deloid2017preparation}. For imaging, the Ag nanowires were sprayed onto a silicon wafer. Secondary electron images of the particles were obtained on a Zeiss ORION NanoFab equipped with a SIMS add-on \cite{wirtz2015high, wirtz2019imaging} using the helium ion beam at an impact energy of 25 keV and a beam current of 0.5 pA. The images have a size of 2048$\times$2048 px with a pixel scale of 1.0309 px/nm.

\subsection*{Scene and synthetic image generation process}
For the synthetic data generation (Fig.~\ref{fig:workflow}b), the open-source render software Blender \cite{blender2018} (Version 2.79.7) is used. In a first step, virtual template particles (see Supplementary Figs.~\ref{supp_fig:sio2_template_particles} and \ref{supp_fig:tio2_template_particles} for SiO$_{2}$ and TiO$_{2}$, respectively) are modelled based on the particle features in the real HIM images. Additionally, a template Blender scene is manually created that contains the substrate as well as the light source. Afterwards, the parameters for the light source and the particle shaders are set manually to achieve a photo-realistic appearance of the particles in the rendered images. Once the correct parameters are set, the python application programming interface (API) of Blender is used to automatically generate a user-defined number of scenes filled with randomly distributed, duplicated and scaled template particles (more details are provided in Supplementary Figs.~\ref{supp_fig:sio2_render_distibution_dirt} and \ref{supp_fig:tio2_render_distibution_dirt} for SiO$_{2}$ and TiO$_{2}$, respectively). To cover the substrate surface impurity, which is present in some HIM images, a random dirt texture is generated for each scene. The dirt texture creation is based on the diamond-square algorithm \cite{fournier1982computer}, also known as the `random midpoint displacement fractal'. After all scenes are generated, each scene is rendered twice in an automated manner. While the first rendering process computes a photo-realistic HIM image, the shaders for the substrate and particles are changed for the second rendering process to obtain a error-free ground truth label. Additionally, aspects of domain randomization \cite{tobin2017domain} were applied by randomly varying the brightness of each scene's light source in the photo-realistic renders (see Supplementary Figs.~\ref{supp_fig:sio2_synthetic} and \ref{supp_fig:tio2_synthetic} for SiO$_{2}$ and TiO$_{2}$, respectively). This way, brighter as well as darker particle images are generated, which increases the variance for the particle appearance in the synthetic data set. An automatically generated 3D scene with the corresponding label scenery is demonstrated in Supplementary Fig.~\ref{supp_fig:sio2_render_scene} and \ref{supp_fig:tio2_render_scene} for SiO$_{2}$ and TiO$_{2}$, respectively. Since the image quality produced by the renderer is adjustable and higher compared to the real HIM images in terms of sharpness, contrast and noise level, we decreased the image quality of the synthetic, photo-realistic images by introducing aliasing effects in combination with additive Gaussian noise. The latter explains the evenly distributed synthetic background images in the t-SNE plots as shown in Fig.~\ref{fig:real_vs_syn_t_SNE}. Aliasing was achieved by upsampling the rendered images from 507$\times$507 px to 2031$\times$2031 px using bilinear interpolation. Also, the synthetic labels were post-processed using binarization with a subsequent erosion (morphological operation) to strengthen the border area between particles.
\\
\\
In summary, for each particle type (SiO$_{2}$, TiO$_{2}$ and Ag) 180 corresponding scenes were generated with their respective synthetic photo-realistic images and labels. The automated process of generating the particle scenes, rendering the photo-realistic images and the labels, as well as the post-processing amounted to approximately 3.5 hours for the SiO$_{2}$ particles, 5.5 hours for the TiO$_{2}$ type and 16 hours for the Ag images. In this context, a Nvidia GeForce GTX 1070 graphics processing unit (GPU) supported the scene renderings while the scene building process was performed by an Intel Xeon W-2102 central processing unit (CPU). The significant increase in time for the Ag images is caused by sophisticated Blender build-in post-processing edge enhancement algorithms in order to achieve accurate contour lines for the virtual objects, especially for the Ag nanowire images (see Supplementary Fig. \ref{supp_fig:ag_synthetic}). The time spent to find and set the correct render settings including the correct light source parameters and shaders is not included in the calculation as it solely depends on the operators' level of experience for the used render software.



\subsection*{Pre-processing and data augmentation}
Prior to the training of the deep CNN, common data augmentation strategies are applied in order to improve the model performance \cite{wang2017effectiveness} by increasing the variance and diversity of the training data. For this purpose, we normalized the pixel intensities within the range of [0, 1] and applied contrast limited adaptive histogram equalization \cite{pizer1987adaptive} (CLAHE) to enhance the contrast of each image in the training data set. Additionally, $90^\circ$ image rotations, flipping, zoom, intensity changes and Gaussian-distributed noise is used.

\subsection*{Convolutional neural network architecture and training}
The U-Nets are implemented and trained on the TensorFlow \cite{abadi2016tensorflow} (version 1.12) framework. We used the original U-Net architecture proposed by Ronneberger et al. \cite{ronneberger2015u} as depicted in Fig. \ref{fig:workflow}c with the same number of feature maps for the encoder and decoder part. Solely the cropping operation is excluded from the skip connections and only one filter kernel is applied in the last 1x1 convolution layer at the end of the decoder part. Additionally, we apply batch normalization \cite{ioffe2015batch} after each convolution layer to reduce overfitting and stabilize the learning process. The weights are initialized with the method proposed by He et al. \cite{he2015delving}. As activation function, the rectified linear unit \cite{glorot2011relu} (ReLU) is used. The training of the network is performed using stochastic gradient descent (SGD) with a patch size of 400$\times$400 px and a mini-batch size of 2. Each image in the training batch was generated by sub-sampling a randomly chosen image from the synthetic data set. Prior to feeding the mini-batch into the network, data augmentation was applied as described in the previous subsection. The number of iterations is set to 150. U-Net\textsubscript{sim} was trained on 180 synthetic images for 500 epochs. U-Net\textsubscript{sim} was trained in a leave-one-out cross-validation setup on 9 SiO$_{2}$ and 8 TiO$_{2}$ images, respectively. Due to the limited amount of real data available the number of epochs for U-Net\textsubscript{real} is set to 300 and the best network state is chosen according to the minimal validation error. To compute a representative probability map as network output, a pixel-wise sigmoid was applied on its last feature map. The pixel-wise sigmoid is defined as:
\\
\begin{equation}
    \qquad \qquad \qquad \qquad \qquad \qquad \qquad \qquad \qquad p(x) = \frac{1}{1 + e^{-x}}
\end{equation}
where $x$ denotes the pixel intensity. The binary segmentation mask is extracted with a threshold of 0.51 on the probability map. In other words, pixels that have a probability of more than 51\% are classified as particle pixels. As loss function cross-entropy loss was applied. We trained the networks (U-Net\textsubscript{real} and U-Net\textsubscript{sim}) with the Adam \cite{kingma2014adam} optimization algorithm using a constant learning rate of 0.001 and the default optimizer parameters in TensorFlow. For each particle type (SiO$_{2}$, TiO$_{2}$ and Ag), a separate U-Net model was trained on a Nvidia GeForce GTX 1070 GPU.

\subsection*{Data set comparison and visualization}
The dimensionality reduction using t-SNE as illustrated in Fig.~\ref{fig:real_vs_syn_t_SNE}c-d is based on features extracted by a VGG16 \cite{simonyan2014very}, a deep CNN, which is implemented in the PyTorch framework \cite{paszke2017automatic} and pre-trained on the ImageNet \cite{JiaDeng2009} data set. The ImageNet is a large data set for object detection tasks containing over 14 million images with more than 2000 categories. Since the VGG16 model is pre-trained on the ImageNet and achieves very high object detection accuracies\cite{simonyan2014very}, it is assumed that the model has learned to extract relevant features to distinguish between various data sets. For data set comparison, image patches of the size of 144$\times$144~px were processed by the VGG16 network, such that real and synthetic SiO$_{2}$ and TiO$_{2}$ images can be represented as data points after the dimensionality reduction. The synthetic data points are generated by randomly sampling from the synthetic images, whereas the data points for real HIM images are extracted sequentially. This is due to the small number of real HIM images (SiO$_{2}$: 9, TiO$_{2}$: 8) compared to the high number of synthetic images (SiO$_{2}$: 180, TiO$_{2}$: 180). Subsequently, a principal component analysis (PCA) \cite{wold1987principal} is applied to reduce the dimensionality of the extracted feature vector from 1000 to 33 components while covering 90\% of its variance. Afterwards, t-SNE \cite{maaten2008tsne} is used to further reduce the dimensionality and to facilitate visualization. This results in 3528 data points for SiO$_{2}$ particles (1764 image patches per data set) and 3136 data points for the TiO$_{2}$ particles (1568 image patches for the real and synthetic data set). Both PCA and the t-SNE algorithm were applied using the scikit-learn \cite{scikit-learn} machine learning python module. 

\subsection*{Model selection}
In order to facilitate the understanding of the model selection step in our workflow (Fig.~\ref{fig:workflow}d), the commonly used approach for selecting an appropriate model in machine learning is explained in the following. When a machine learning algorithm, i.e. a deep CNN, is trained in a supervised fashion to solve a specific task, in general three stages must be passed in order to select the best model: training, validation and testing. While training, the algorithm fits on the training data set in order to perform a specific task (i.e. segmentation). Specifically, during this phase the algorithm iteratively tunes its internal parameters, the so-called `weights' (weighted connections between neurons) in an artificial neural network, after each training step to improve its decision-making. In this context, a machine learning algorithm that is trained on a data set is termed `model'. To prevent the model from overfitting on the training set, the validation data is used to assess the model performance after each training iteration using a pre-defined evaluation metric - the so-called `validation loss'. The model, which achieves the minimal loss on the validation set is selected as the best model and is evaluated afterwards on the previously unseen test data.
\\
\\
For the manual model selection, we conducted a user study with 7 participants, consisting of three experts on segmentation and four non-experts. We trained a U-Net for 500 epochs on synthetic SiO$_{2}$ and TiO$_{2}$ data, respectively. For both particle types, a representative real HIM validation image was selected and manually annotated. The validation images for SiO$_{2}$ and TiO$_{2}$ as well as the corresponding manual GT annotations are visualized in Fig.~\ref{fig:study_evaluation_validation}c. After training, each participant was asked to choose five models out of 500 (Model 1 to Model $n_\text{epochs}$ in Fig.~\ref{fig:workflow}d) that performed best based on the visual evaluation of the model's segmentation performances on the real validation image.

\subsection*{Post-processing}
The segmentation output of the U-Nets (U-Net\textsubscript{real} and U-Net\textsubscript{sim}) is post-processed in order to enhance the segmentation quality. For this purpose, we utilized area opening on the binary segmentation mask with an area size of 400 px for the SiO$_{2}$ and 600 px for the TiO$_{2}$ images to remove noise that falls below a certain area size. Subsequently, we apply a distance transform watershed \cite{Soille1990DeterminingWI, legland2016morpholibj} algorithm using the MorphoLibJ \cite{legland2016morpholibj} plugin of the open-source ImageJ/Fiji \cite{schindelin2012fiji} software to further separate particles with touching borders. In this context, we used the default settings with a dynamic of 20 for the TiO$_{2}$ particles and a normalized output combined with a dynamic of 4 for the SiO$_{2}$ images.

\subsection*{Metrics}
For the quantitative evaluation several measures are utilized, including accuracy, precision, recall and the Dice similarity index \cite{dice1945measures}, also known as F1 score. The metrics are defined as:
\begin{equation}
    Accuracy = \frac{TP + TN}{TP + FP + TN + FN} \quad Precision = \frac{TP}{TP + FP} \quad Recall = \frac{TP}{TP + FN} \quad F1 = \frac{2 \cdot TP} {2 \cdot TP + FP + FN}
    \label{eq:test}
\end{equation}
True positives (TP) and true negatives (TN) denote pixels, that are classified correctly as particles or background, respectively. False positives (FP) and false negatives (FN) are misclassified pixels that do not appear in the manual ground truth segmentation. The values for the measures vary from 0 to 1, while a value of 0 represents the worst possible accuracy whereas 1 denotes a perfect segmentation result. In literature \cite{ghafoorian2017location}, a F1 score of 0.7 or higher is already considered as a good segmentation.

\section*{Acknowledgements}
The research leading to these results has received funding from the European Union's Horizon 2020 Research and innovation program (Grant agreement No. 720964) as well as from the European Research Council (ERC) (Grant agreement No. 810316).

\section*{Author contributions}
L.M. conceived the idea and designed and implemented the workflow together with D.W.. L.K., N.G., P.P., F.V., A.I., A.M. and S.C. contributed to the design of the data simulation and the workflow. T.W., J.N.A. and P.P defined the needs for image segmentation. J.N.A and O.D.C recorded the SE images on the HIM. I.N. contributed to the writing of the introduction section and commented on the manuscript. Y.H. and C.J. helped with the evaluation of the results. L.M. conducted the simulations, implemented and trained the deep learning models, and evaluated the results. L.M., N.G., L.K., F.V., C.J., I.N., Y.H., A.M. and S.C. discussed the results and suggested improvements. L.M. and S.C. iterated the first version of the manuscript and all authors contributed to the final text. 

\section*{Competing interests}
The authors declare no competing interests.

\clearpage

\section*{Supplementary Information}
\setcounter{figure}{0}
\makeatletter 
\renewcommand{\thefigure}{S\@arabic\c@figure}
\makeatother
\renewcommand{\thesection}{\Roman{section}} 
\renewcommand{\thesubsection}{\thesection.\Roman{subsection}}
\section{Helium Ion Microscopy images}
\subsection{SiO\texorpdfstring{\textsubscript{2}}{2}~nanoparticles}
\begin{figure}[h!]
\centering
\includegraphics[scale=0.95]{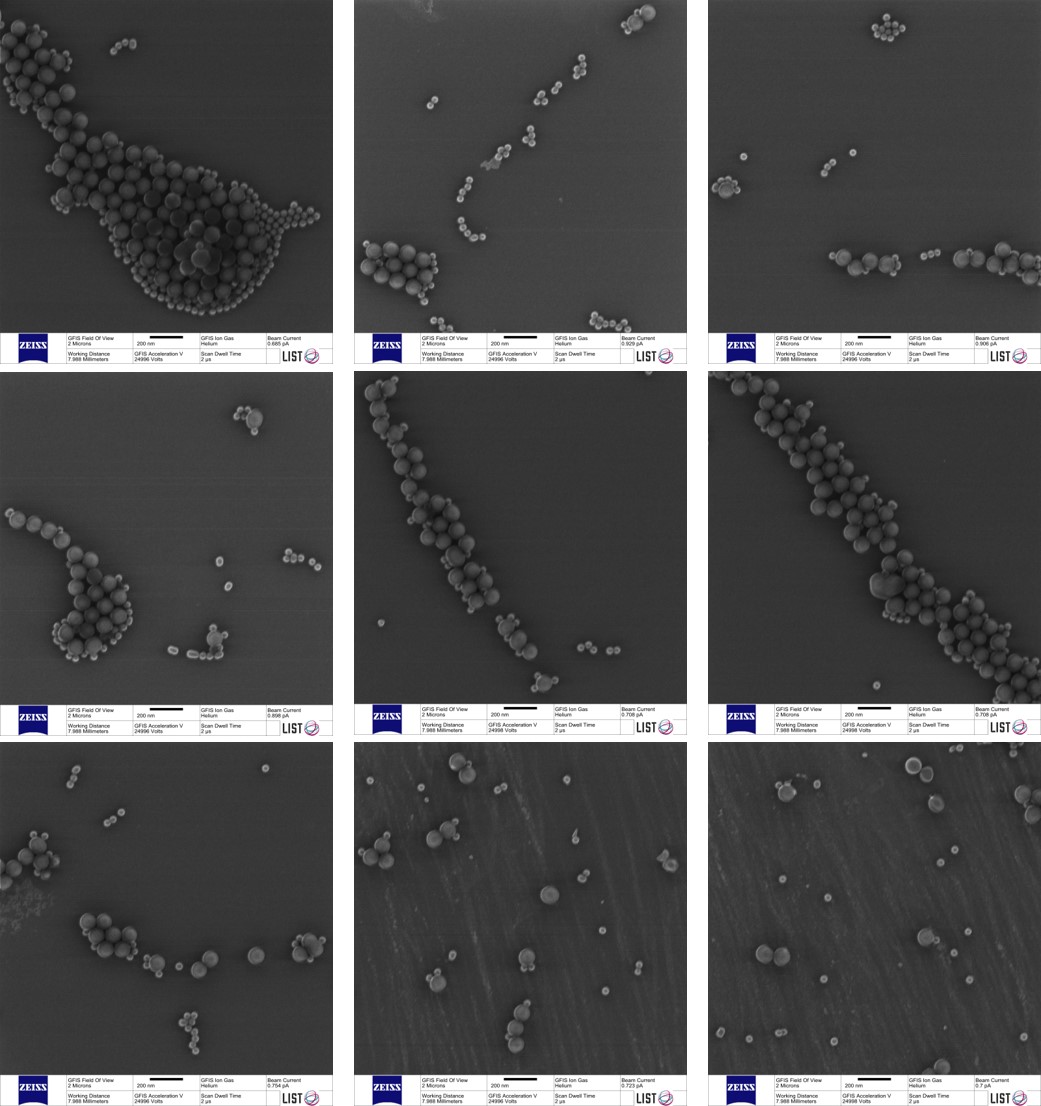}
\caption{\textbf{HIM images of SiO$_{2}$ nanoparticles.} Note the substrate surface impurity which is present in the last two images of the bottom row.}
\label{supp_fig:sio2_raw}
\end{figure}

\clearpage
\subsection{TiO\texorpdfstring{\textsubscript{2}}{2}~nanoparticles nanoparticles}
\begin{figure}[h!]
\centering
\includegraphics[scale=0.95]{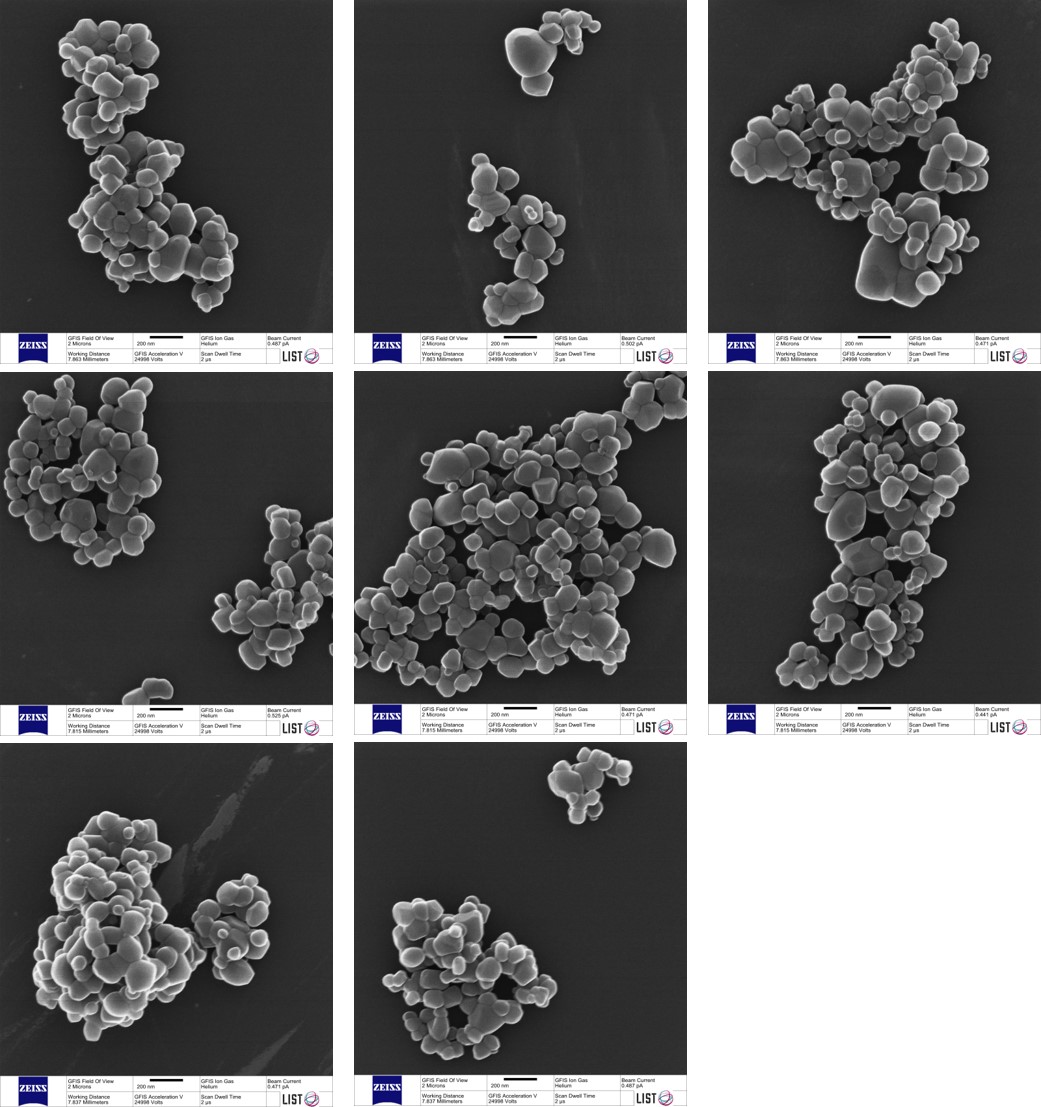}
\caption{\textbf{HIM images of TiO$_{2}$ nanoparticles.}}
\label{supp_fig:tio2_raw}
\end{figure}

\clearpage
\subsection{Ag nanoparticles}
\begin{figure}[h!]
\centering
\includegraphics[scale=0.95]{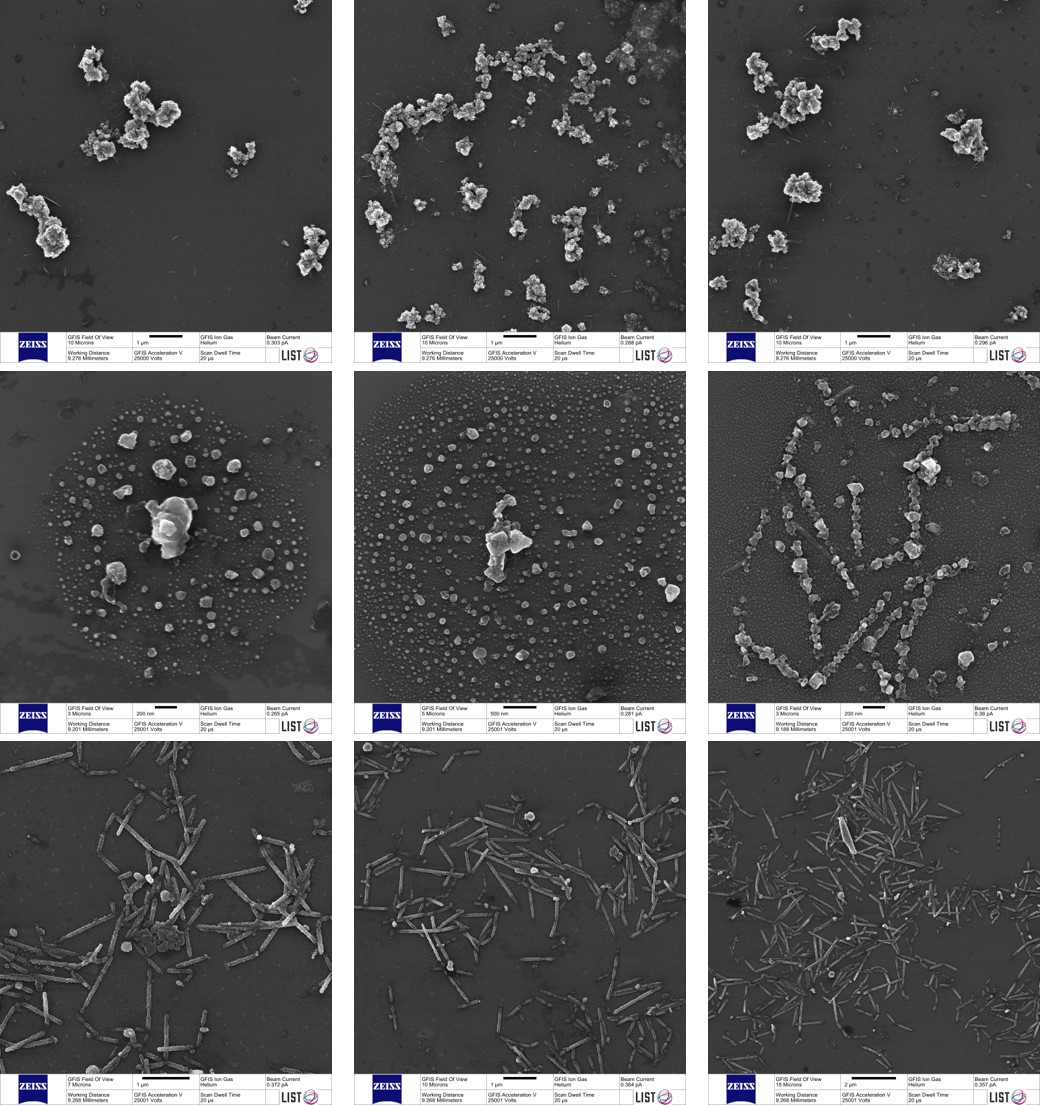}
\caption{\textbf{HIM images of Ag particles, nanorods and nanowires.}}
\label{supp_fig:ag_raw}
\end{figure}
\clearpage
\section{Additional information on synthetic image generation}
\subsection{SiO\texorpdfstring{\textsubscript{2}}{2}}
\begin{figure}[h!]
\centering
\includegraphics[scale=1]{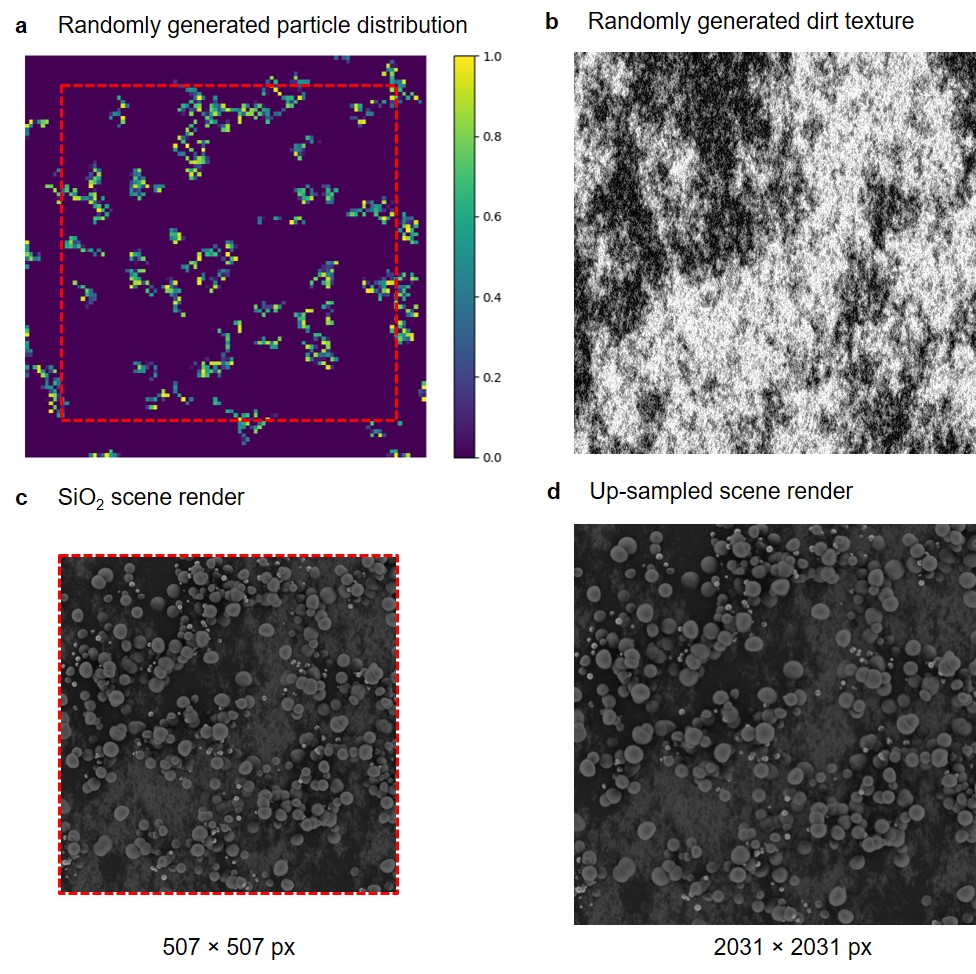}
\caption{\textbf{Particle center coordinate distribution, impurity texture and scene render with post-processing. a},  Randomly generated distribution map for particle centers containing 989 particles. Each pixel represents a coordinate (x and y position) of a single particle center in the virtual scene. The normalized pixel intensities denote the z-coordinates (heights). The red dotted rectangle represents the perception field of the camera by setting a random zoom factor before rendering the scene. For each scene, a new distribution map and a new camera zoom factor are computed. By this way, various particle sizes are achieved. \textbf{b}, Randomly generated dirt texture to simulate substrate surface impurity. \textbf{c}, Raw scene render. \textbf{d}, Up-sampled scene render from $507 \times 507$ px to $2031 \times 2031$ px using bilinear interpolation to generate similar blurry particle contours as of the real SiO$_{2}$ HIM particle images. Additionally, Gaussian noise is added. }
\label{supp_fig:sio2_render_distibution_dirt}
\end{figure}

\clearpage
\begin{figure}[h!]
\centering
\includegraphics[scale=1]{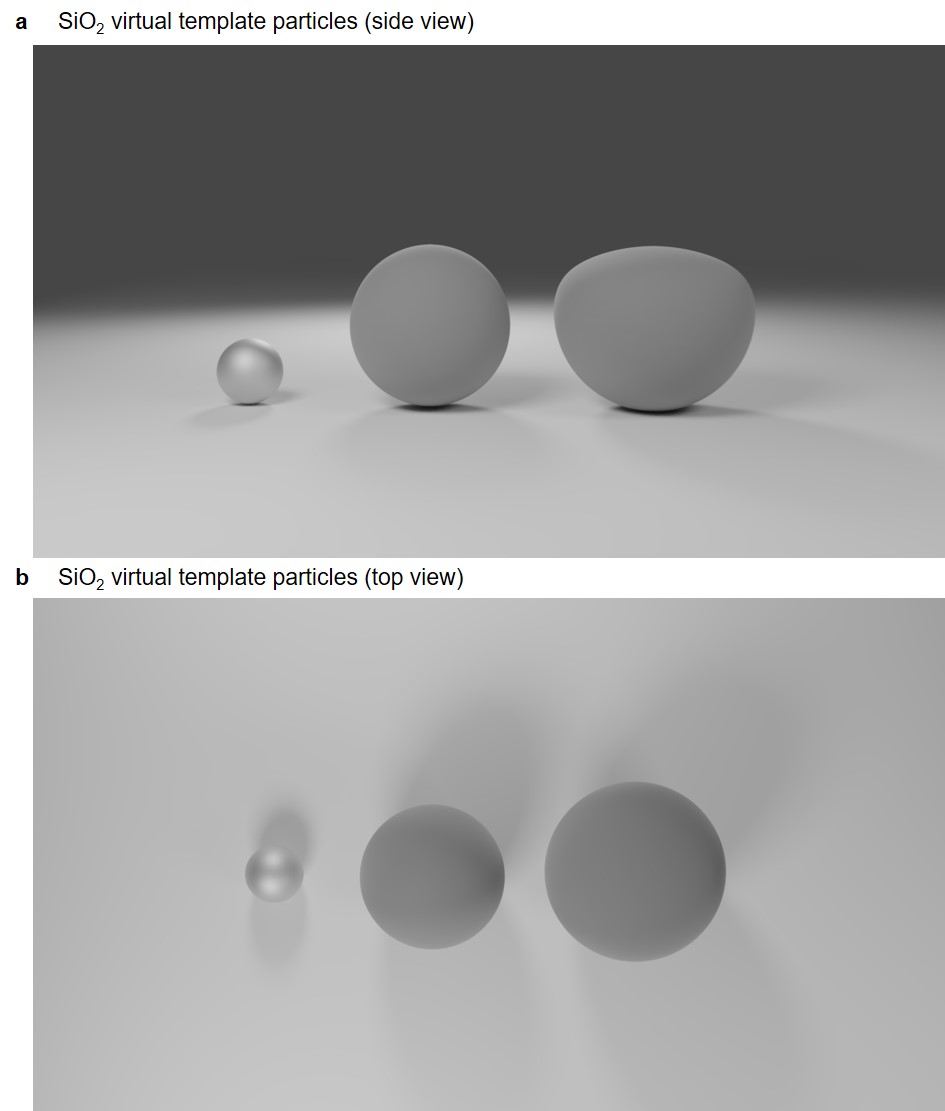}
\caption{\textbf{Manually designed virtual SiO$_{2}$ template particles.} Three spherical particles were designed to imitate real SiO$_{2}$ particles. The two big spheres have a diffuse shader, while for the smaller one a glossy shader was used to mimic a brighter appearance in the synthetic images.}
\label{supp_fig:sio2_template_particles}
\end{figure}

\clearpage
\begin{figure}[h!]
\centering
\includegraphics[scale=1]{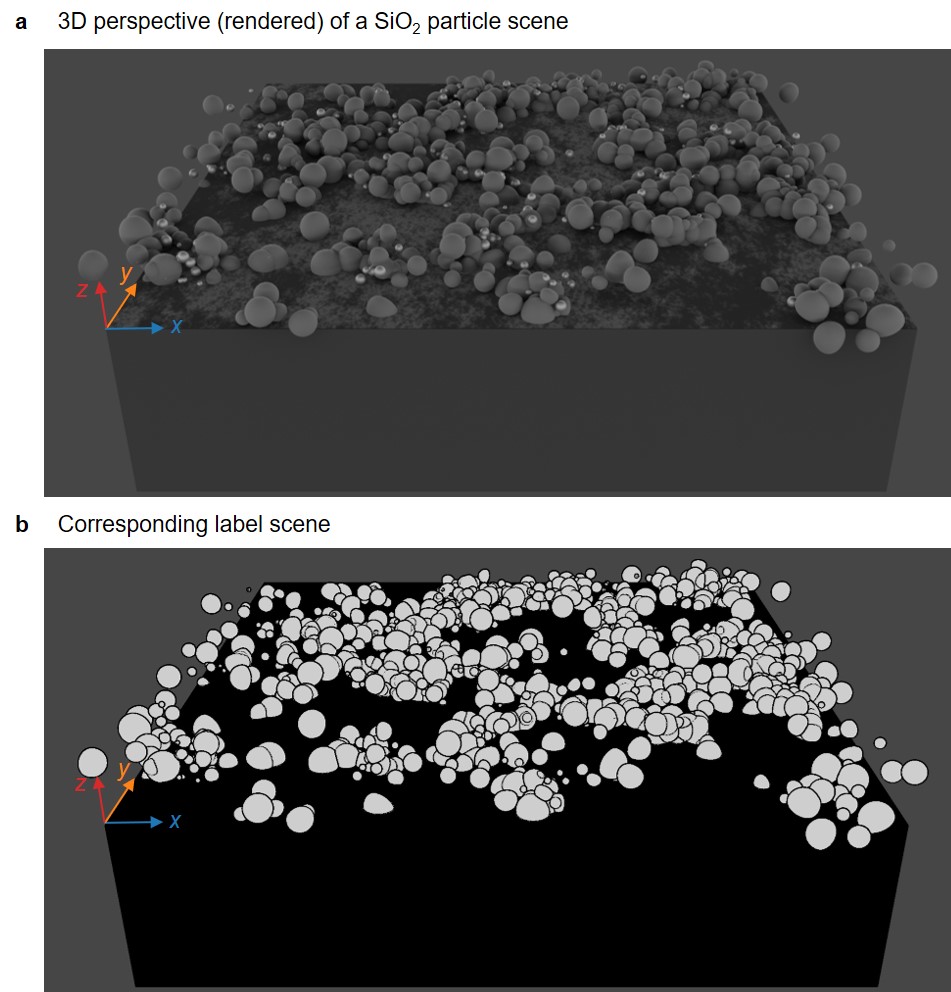}
\caption{\textbf{Virtual SiO$_{2}$ particle scenes. a}, Rendered 3D perspective of a SiO$_{2}$ scene. The bottom left corner of the virtual substrate represents the origin of coordinates, which is used as reference when particles are loaded into the scene according to a distribution map. \textbf{b}, Corresponding 3D label render, which is achieved by changing the shaders of the substrate and the particles. Additionally, a Blender in-build post-processing pipeline (Compositor) is used to obtain precise particle contours.}
\label{supp_fig:sio2_render_scene}
\end{figure}


\clearpage
\subsection{TiO\texorpdfstring{\textsubscript{2}}{2}}
\begin{figure}[h!]
\centering
\includegraphics[scale=1]{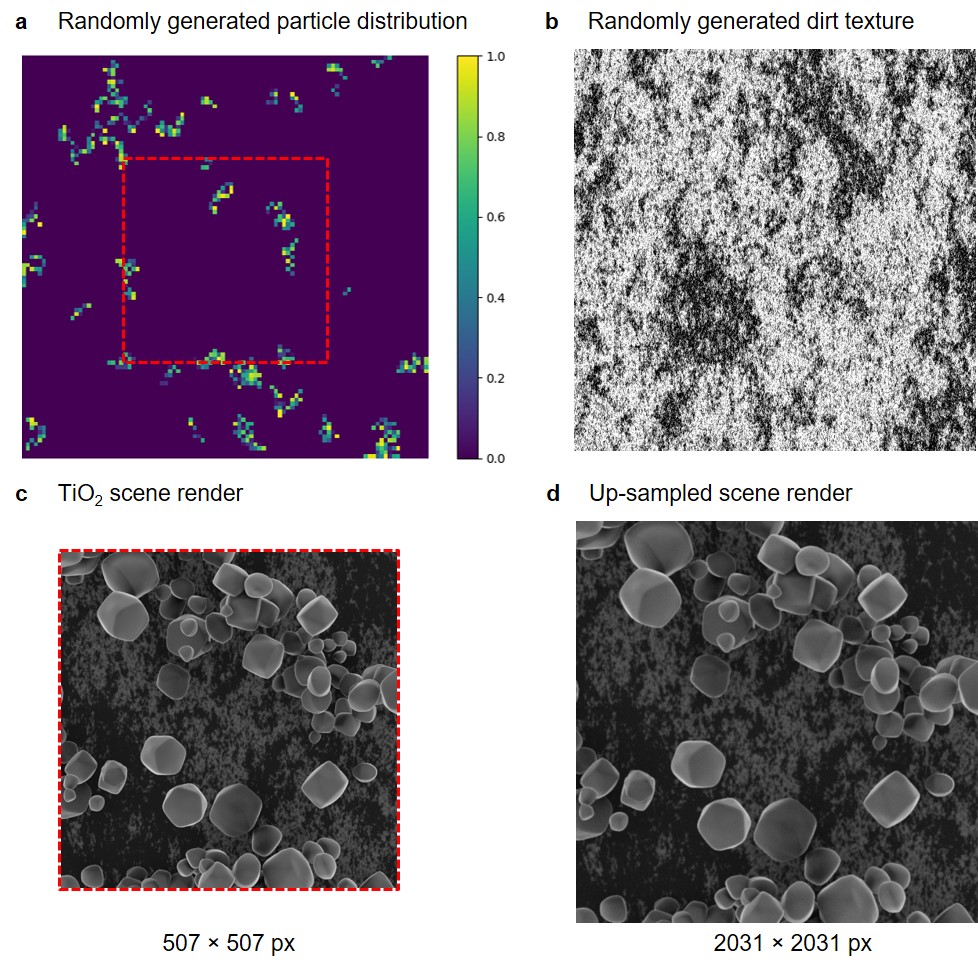}
\caption{\textbf{Particle center coordinate distribution, impurity texture and scene render with post-processing. a},  Randomly generated distribution map for particle centers containing 539 particles. Each pixel represents a coordinate (x and y position) of a single particle center in the virtual scene. The normalized pixel intensities denote the z-coordinates (heights). The red dotted rectangle represents the perception field of the camera by setting a random zoom factor before rendering the scene. For each scene, a new distribution map and a new camera zoom factor are computed. By this way, various particle sizes are achieved. \textbf{b}, Randomly generated dirt texture to simulate substrate surface impurity. \textbf{c}, Raw scene render. \textbf{d}, Up-sampled scene render from $507 \times 507$ px to $2031 \times 2031$ px using bilinear interpolation to generate similar blurry particle contours as of the real TiO$_{2}$ HIM particle images. Additionally, Gaussian noise is added. }
\label{supp_fig:tio2_render_distibution_dirt}
\end{figure}

\clearpage
\begin{figure}[h!]
\centering
\includegraphics[scale=1]{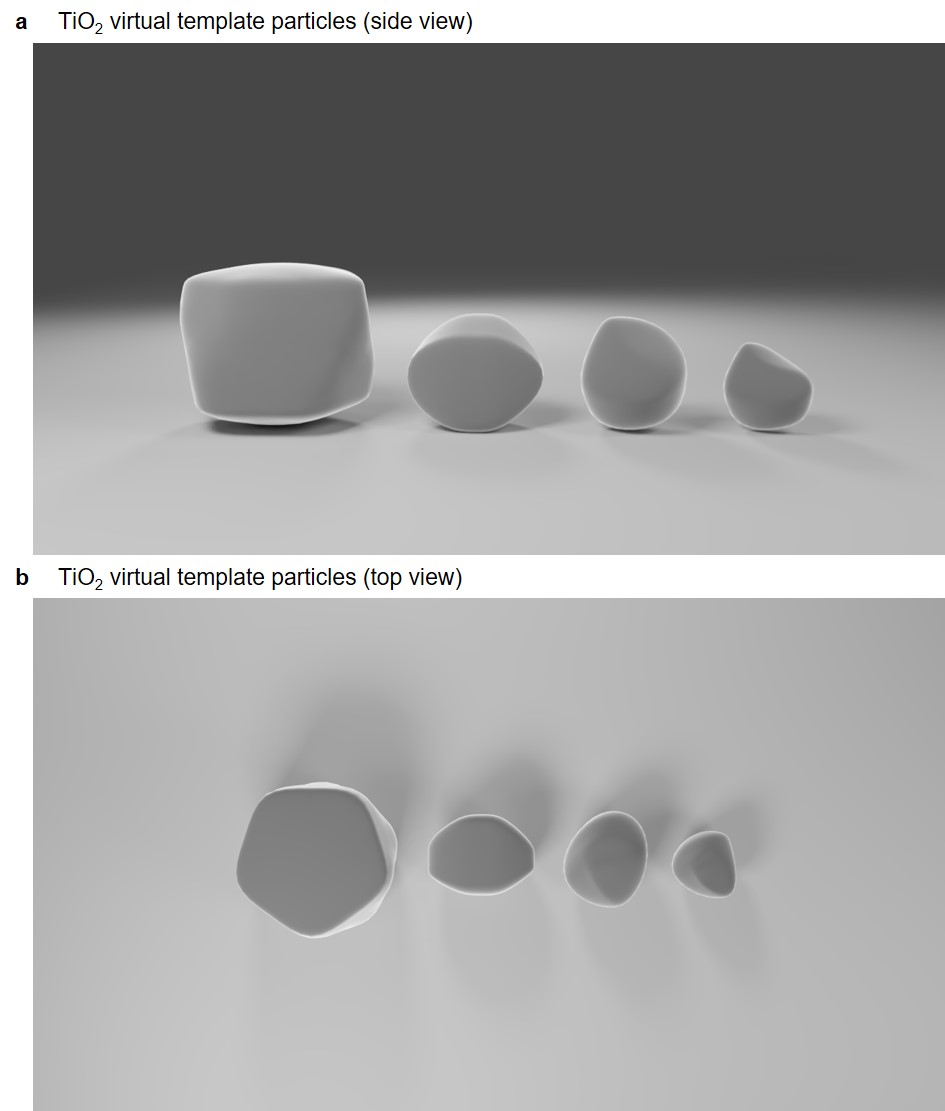}
\caption{\textbf{Manually designed virtual TiO$_{2}$ template particles.} Four particles were designed to imitate real TiO$_{2}$ particles. For all TiO$_{2}$ template particles the same shader was used that mimics the edge-effect, which is characteristic for certain materials in HIM or SEM images.}
\label{supp_fig:tio2_template_particles}
\end{figure}

\clearpage
\begin{figure}[h!]
\centering
\includegraphics[scale=1]{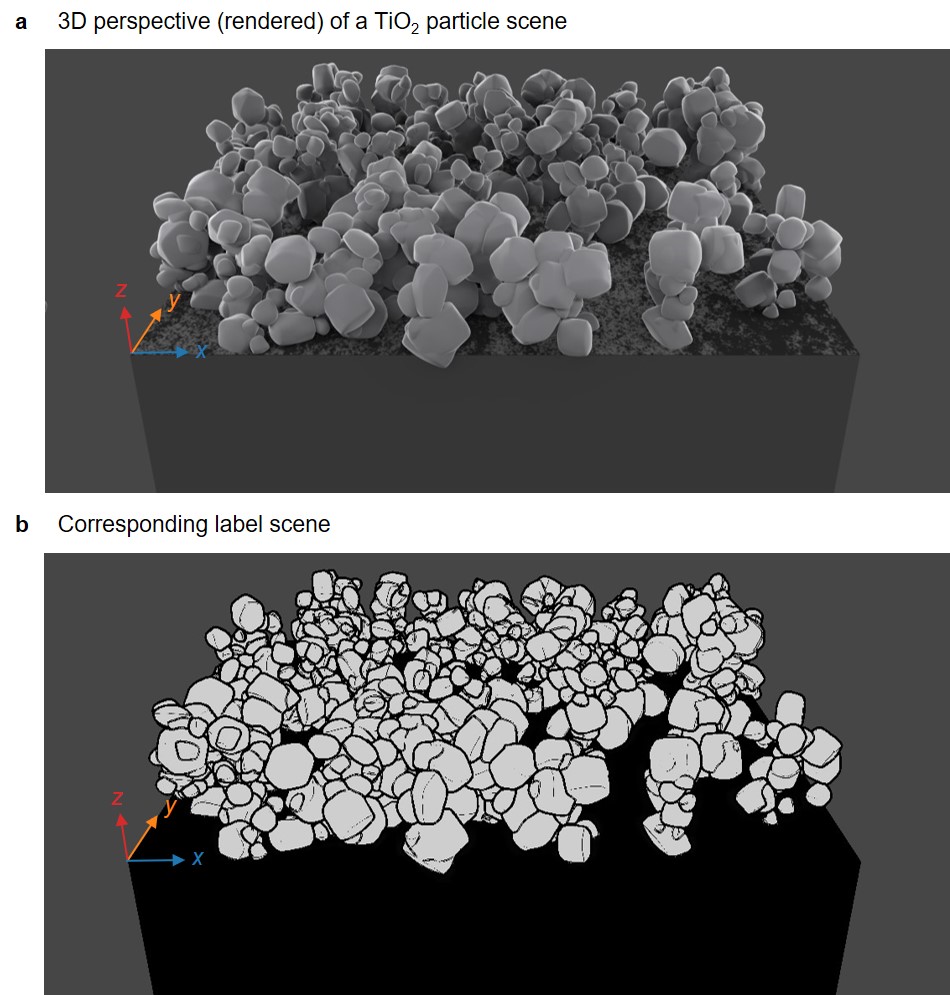}
\caption{\textbf{Virtual TiO$_{2}$ particle scenes. a}, Rendered 3D perspective of a TiO$_{2}$ scene. The bottom left corner of the virtual substrate represents the origin of coordinates, which is used as reference when particles are loaded into the scene according to a distribution map. \textbf{b}, Corresponding 3D label render, which is achieved by changing the shaders of the substrate and the particles. Additionally, a Blender in-build post-processing pipeline (Compositor) is used to obtain precise particle contours.}
\label{supp_fig:tio2_render_scene}
\end{figure}


\clearpage
\section{Synthetic HIM images}
\subsection{SiO\texorpdfstring{\textsubscript{2}}{2} nanoparticles}
\begin{figure}[h!]
\centering
\includegraphics[scale=0.9]{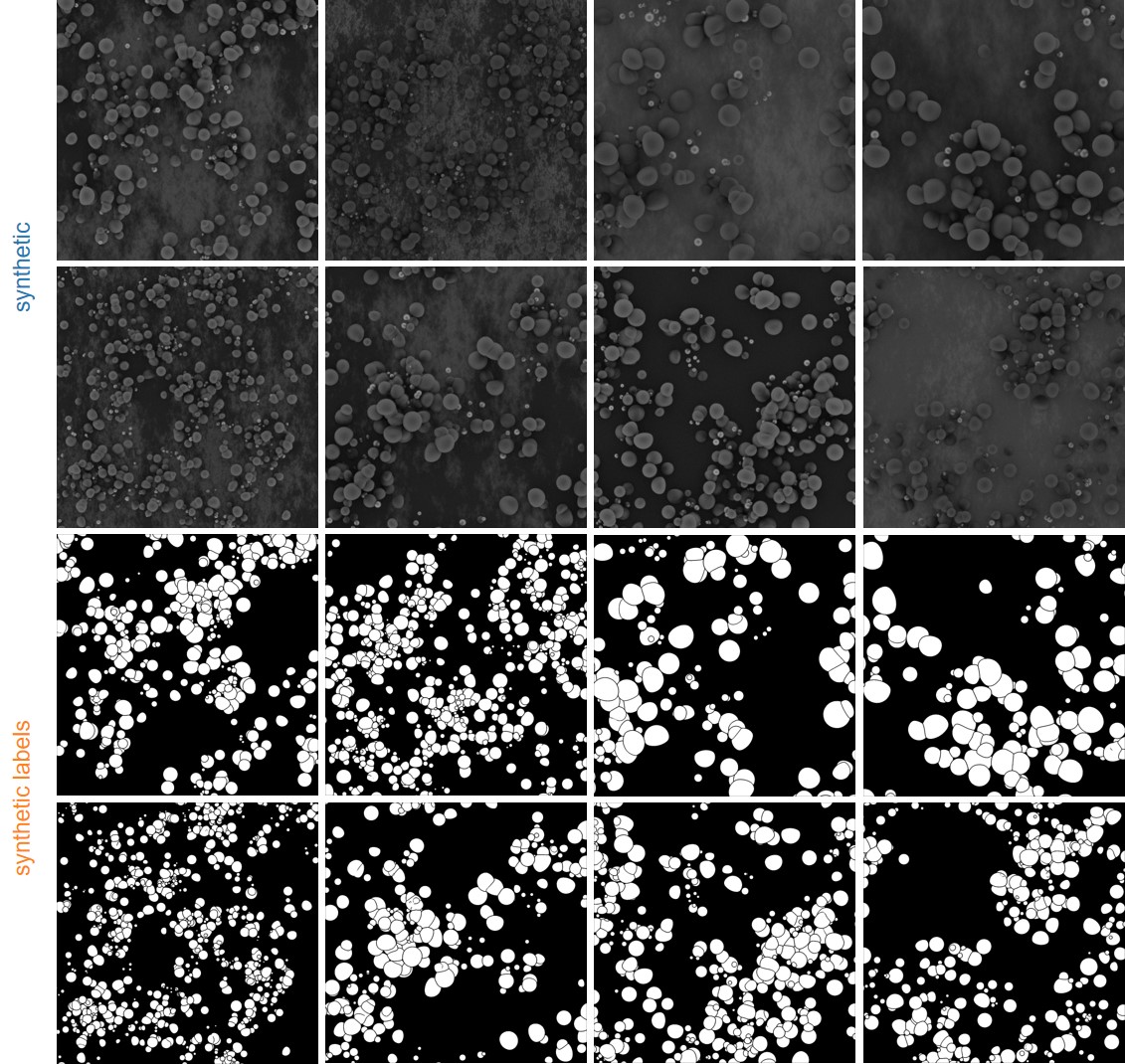}
\caption{\textbf{Synthetic photo-realistic SiO$_{2}$ nanoparticles images with the respective labels.}  The images are generated automatically using the data generation pipeline after the render settings (shaders and light source) are set manually. The automatically extracted labels do not contain any mislabeling and are error-free in terms of contour lines.}
\label{supp_fig:sio2_synthetic}
\end{figure}

\clearpage
\subsection{TiO\texorpdfstring{\textsubscript{2}}{2} nanoparticles}
\begin{figure}[h!]
\centering
\includegraphics[scale=0.9]{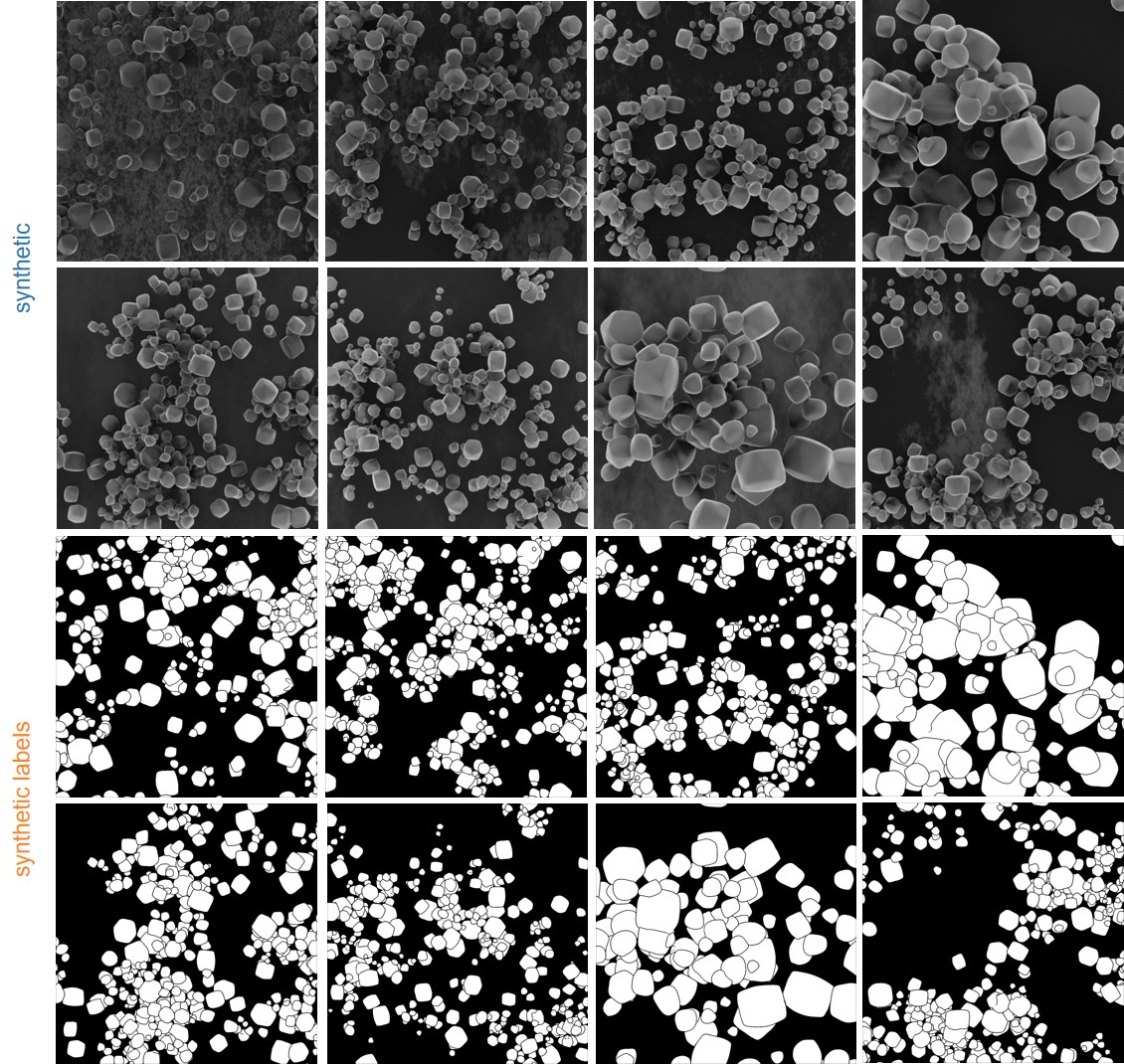}
\caption{\textbf{Synthetic photo-realistic TiO$_{2}$ nanoparticles images with the respective labels.}  The images are generated automatically using the data generation pipeline after the render settings (shaders and light source) are set manually. The automatically extracted labels do not contain any mislabeling and are error-free in terms of contour lines.}
\label{supp_fig:tio2_synthetic}
\end{figure}

\clearpage
\subsection{Ag}
\begin{figure}[h!]
\centering
\includegraphics[scale=0.9]{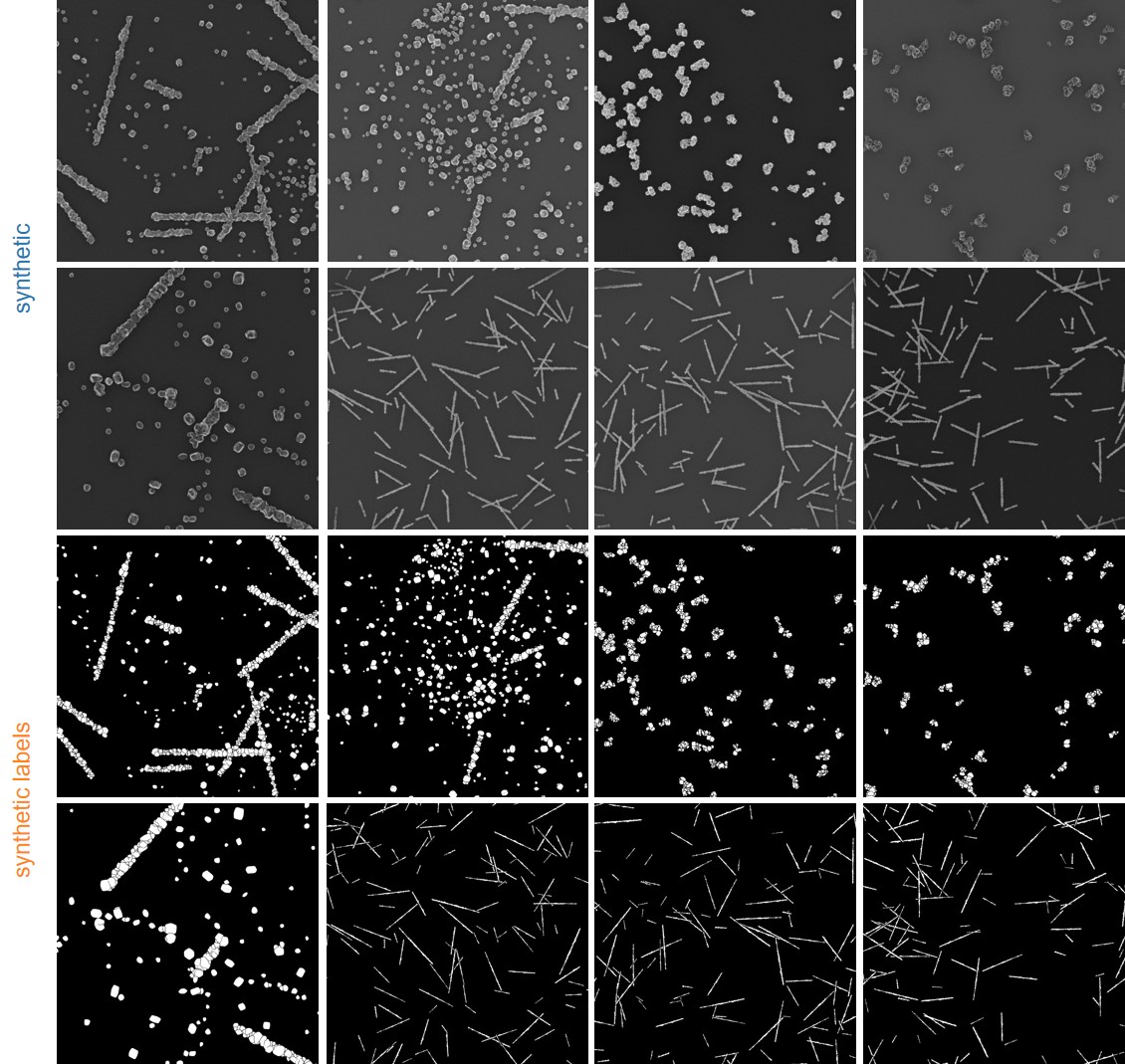}
\caption{\textbf{Synthetic Ag images with the respective labels.}  The images are generated automatically using the data generation pipeline after the render settings (shaders and light source) are set manually. The automatically extracted labels do not contain any mislabeling and are error-free in terms of contour lines. The data set was designed such that in case of a more detailed view of nanorods, the labels contain a separation of the composition the nanord or nanowire is made of (see top left image with its corresponding label).}
\label{supp_fig:ag_synthetic}
\end{figure}


\clearpage
\section{Comparison of segmentation results}
\subsection{SiO\texorpdfstring{\textsubscript{2}}{2} nanoparticles}
\begin{figure}[h!]
\centering
\includegraphics[scale=0.95]{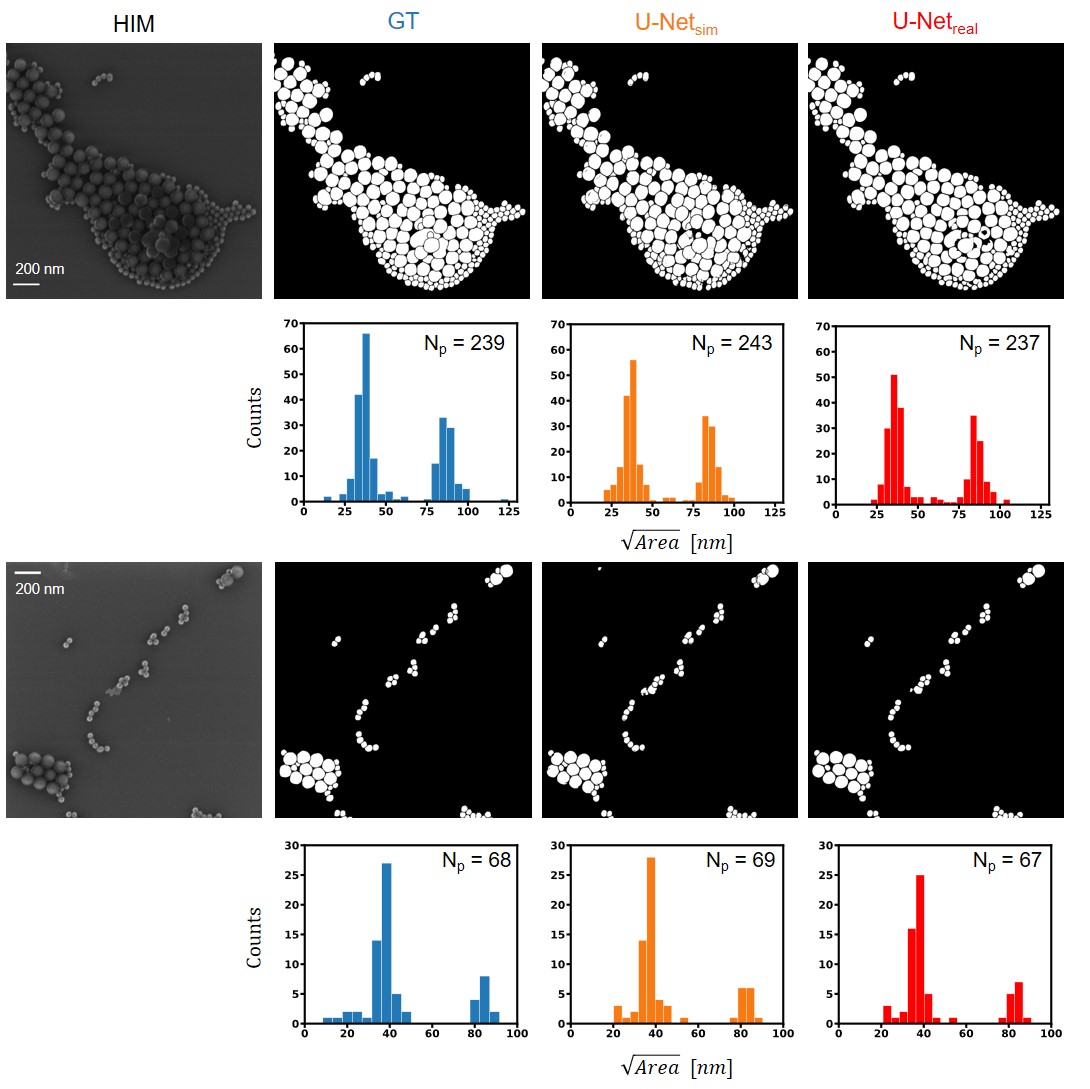}
\caption{\textbf{Post-processed segmentation results of U-Net\textsubscript{sim} and U-Net\textsubscript{real} for HIM SiO$_{2}$ nanoparticle images, compared to the ground truth (GT) annotation}. N\textsubscript{p} denotes the number of particles.}
\label{supp_fig:sio2_histograms_1}
\end{figure}

\clearpage
\begin{figure}[h!]
\centering
\includegraphics[scale=0.95]{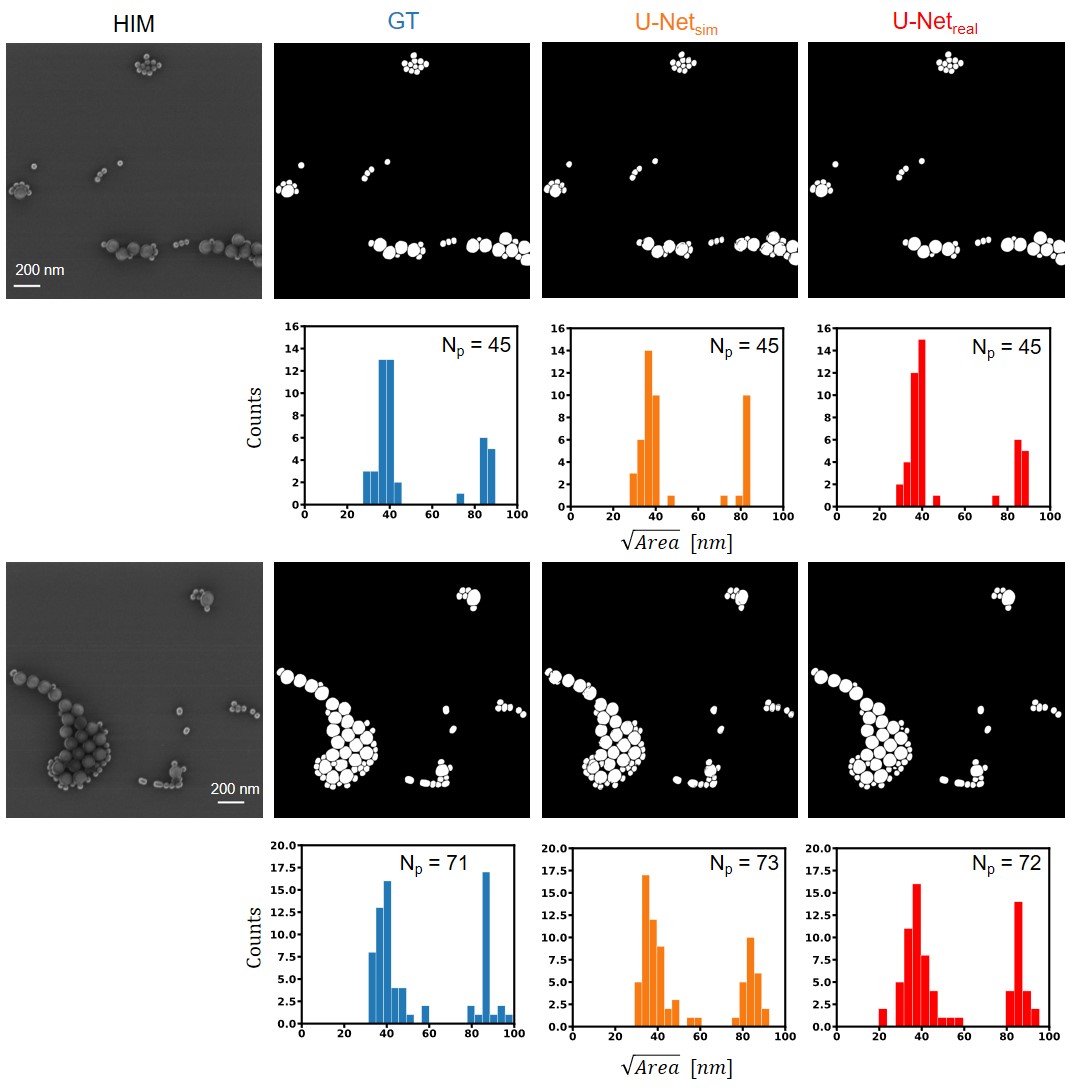}
\caption{\textbf{Post-processed segmentation results of U-Net\textsubscript{sim} and U-Net\textsubscript{real} for HIM SiO$_{2}$ nanoparticle images, compared to the ground truth (GT) annotation}. N\textsubscript{p} denotes the number of particles.}
\label{supp_fig:sio2_histograms_2}
\end{figure}

\clearpage
\begin{figure}[h!]
\centering
\includegraphics[scale=0.95]{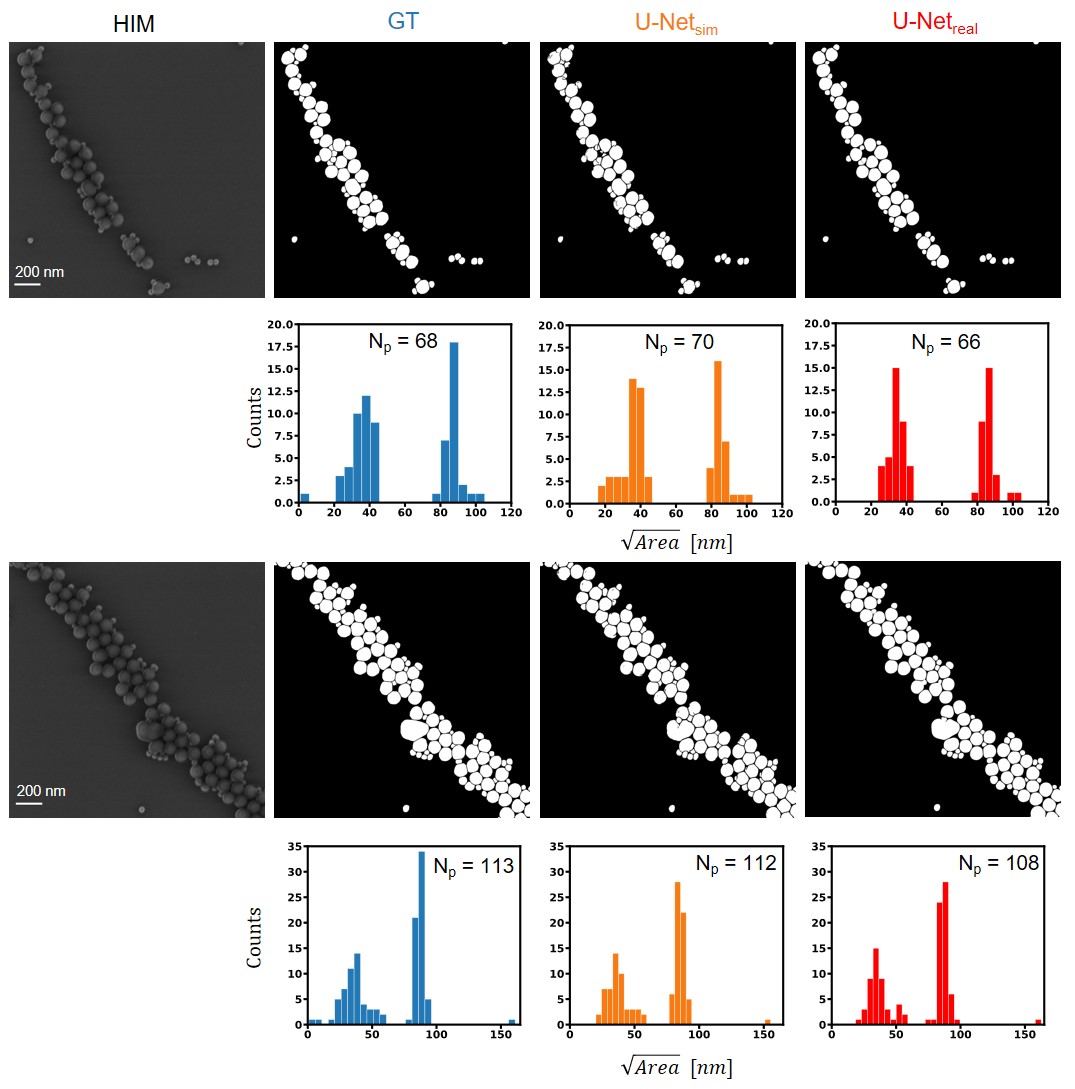}
\caption{\textbf{Post-processed segmentation results of U-Net\textsubscript{sim} and U-Net\textsubscript{real} for HIM SiO$_{2}$ nanoparticle images, compared to the ground truth (GT) annotation}. N\textsubscript{p} denotes the number of particles.}
\label{supp_fig:sio2_histograms_3}
\end{figure}

\clearpage
\begin{figure}[h!]
\centering
\includegraphics[scale=0.95]{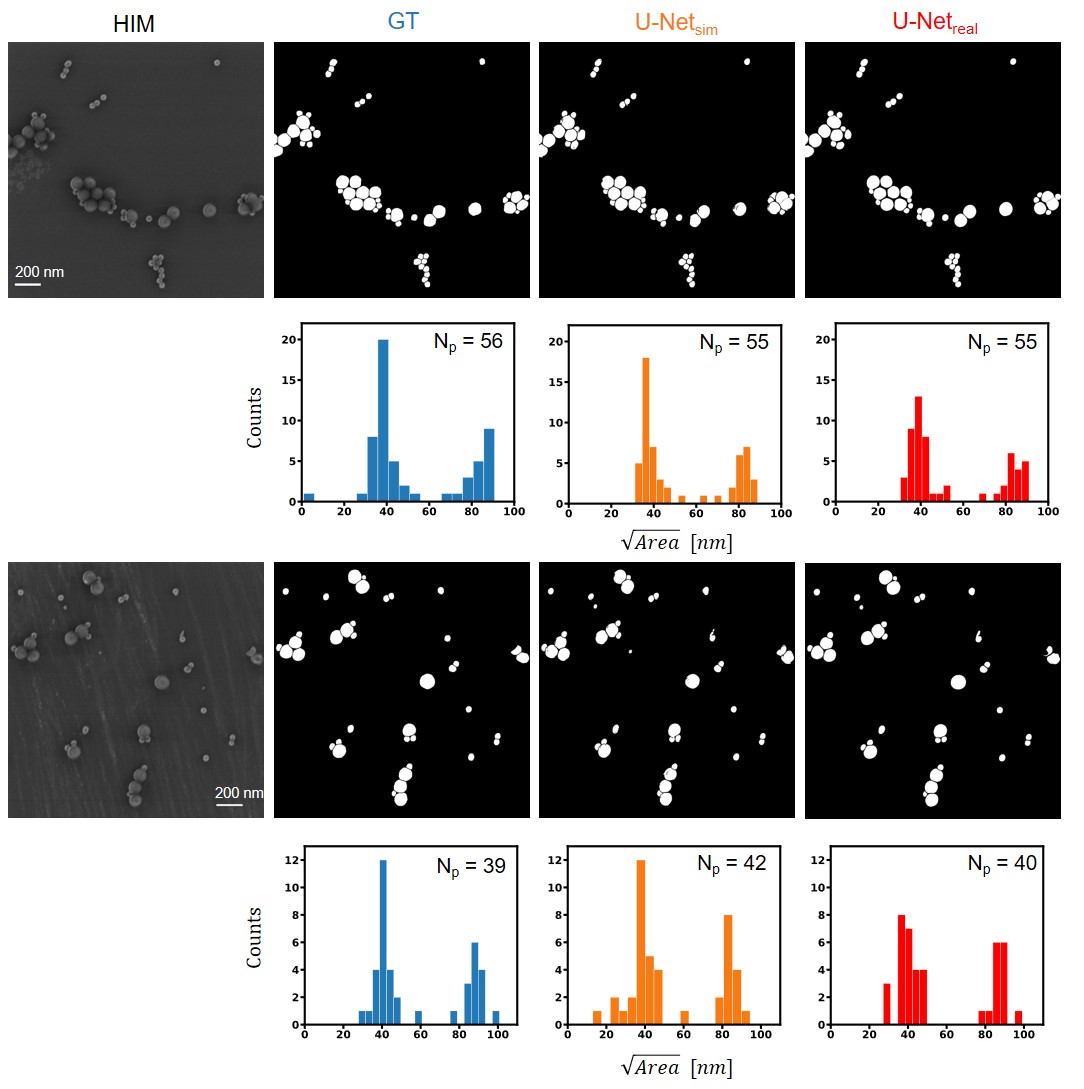}
\caption{\textbf{Post-processed segmentation results of U-Net\textsubscript{sim} and U-Net\textsubscript{real} for HIM SiO$_{2}$ nanoparticle images, compared to the ground truth (GT) annotation}. N\textsubscript{p} denotes the number of particles.}
\label{supp_fig:sio2_histograms_4}
\end{figure}

\clearpage
\begin{figure}[h!]
\centering
\includegraphics[scale=0.95]{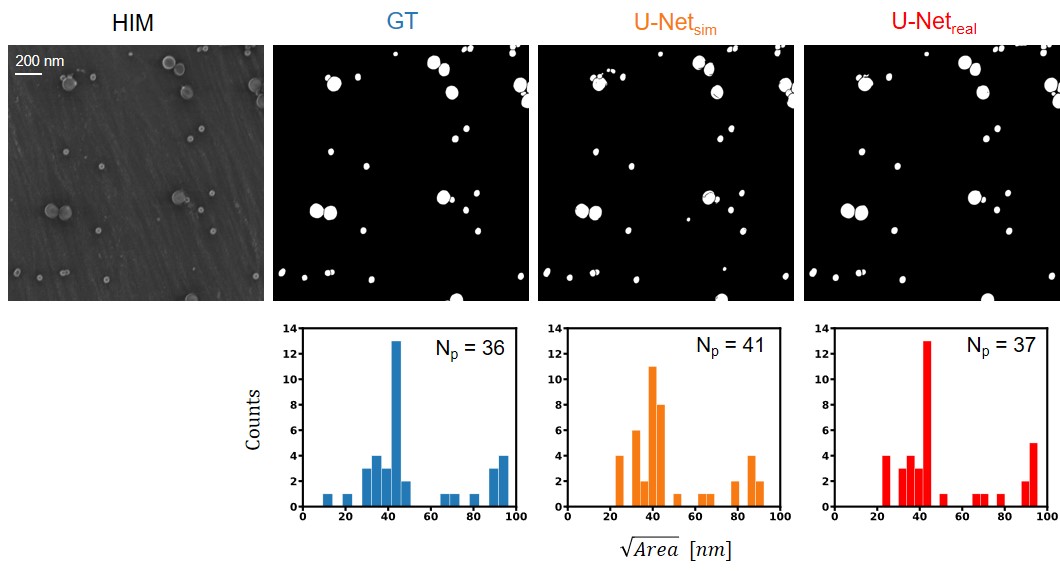}
\caption{\textbf{Post-processed segmentation results of U-Net\textsubscript{sim} and U-Net\textsubscript{real} for HIM SiO$_{2}$ nanoparticle images, compared to the ground truth (GT) annotation}. N\textsubscript{p} denotes the number of particles.}
\label{supp_fig:sio2_histograms_5}
\end{figure}

\clearpage
\subsection{TiO\texorpdfstring{\textsubscript{2}}{2} nanoparticles}
\begin{figure}[h!]
\centering
\includegraphics[scale=0.95]{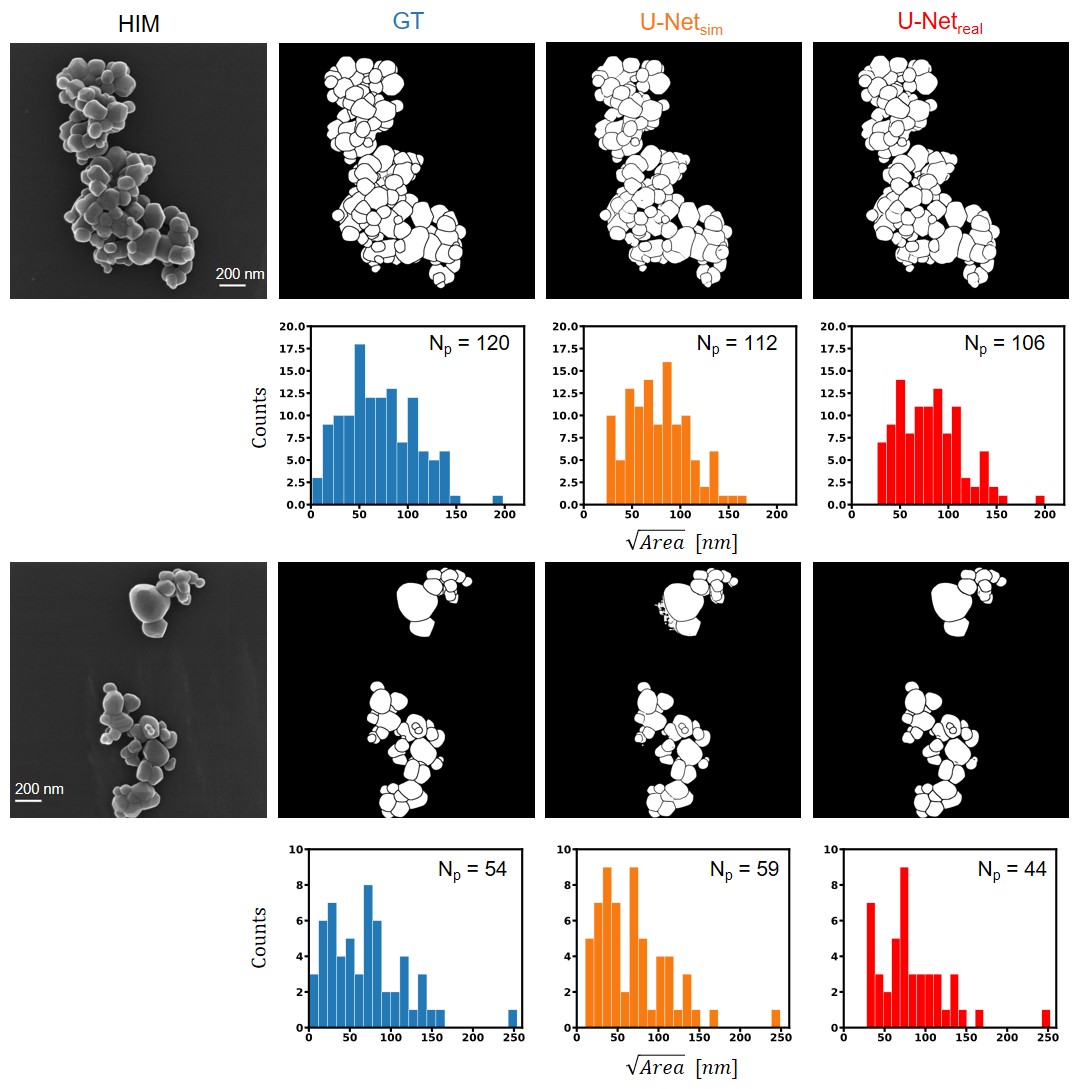}
\caption{\textbf{Post-processed segmentation results of U-Net\textsubscript{sim} and U-Net\textsubscript{real} for HIM SiO$_{2}$ nanoparticle images, compared to the ground truth (GT) annotation}. N\textsubscript{p} denotes the number of particles. Note that due to outlier correction in the post-processing, small objects are removed from the segmentation that fall below a certain size, which is also reflected in the histograms.}
\label{supp_fig:tio2_histograms_1}
\end{figure}

\clearpage
\begin{figure}[h!]
\centering
\includegraphics[scale=0.95]{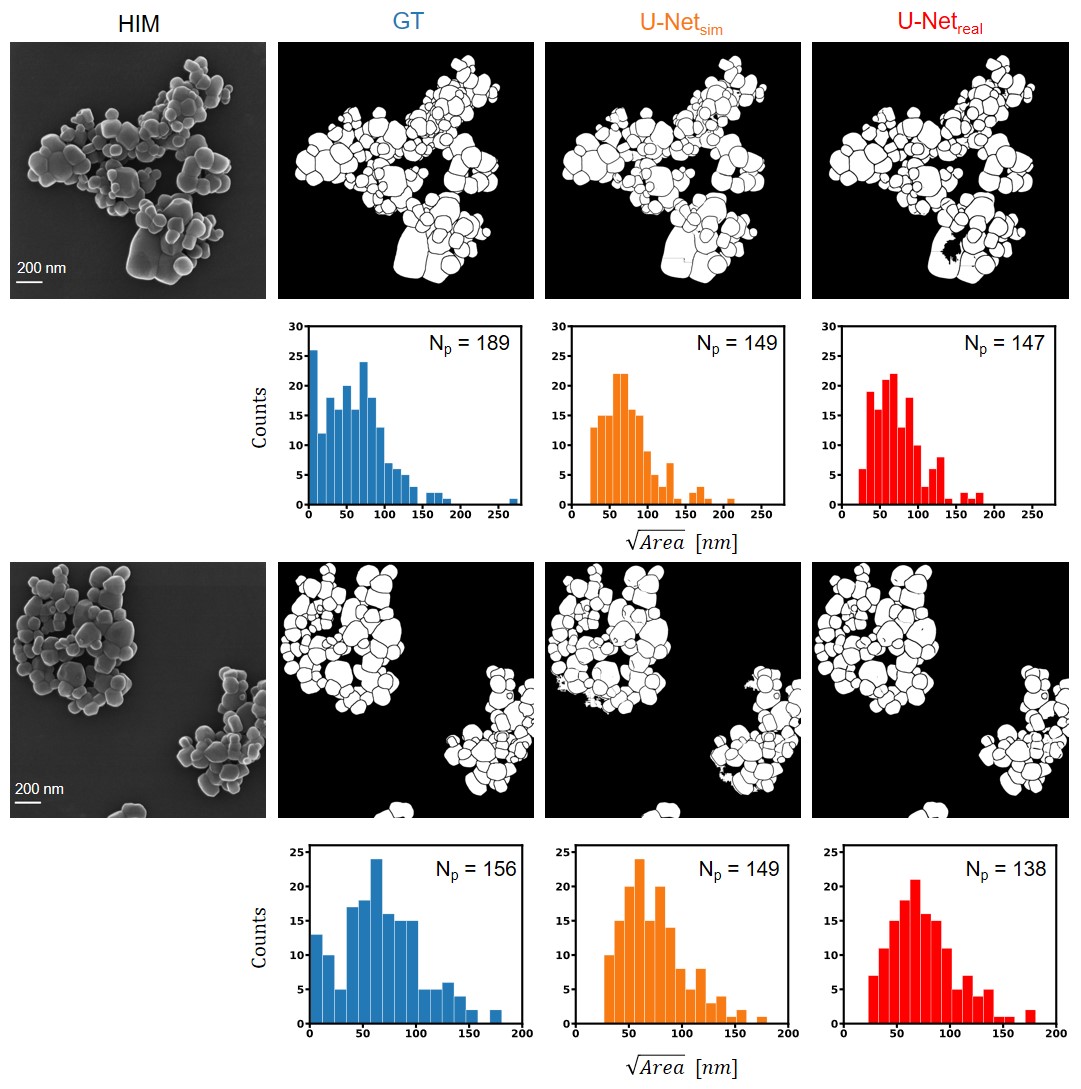}
\caption{\textbf{Post-processed segmentation results of U-Net\textsubscript{sim} and U-Net\textsubscript{real} for HIM SiO$_{2}$ nanoparticle images, compared to the ground truth (GT) annotation}. N\textsubscript{p} denotes the number of particles. Note that due to outlier correction in the post-processing, small objects are removed from the segmentation that fall below a certain size, which is also reflected in the histograms.}
\label{supp_fig:tio2_histograms_2}
\end{figure}

\clearpage
\begin{figure}[h!]
\centering
\includegraphics[scale=0.95]{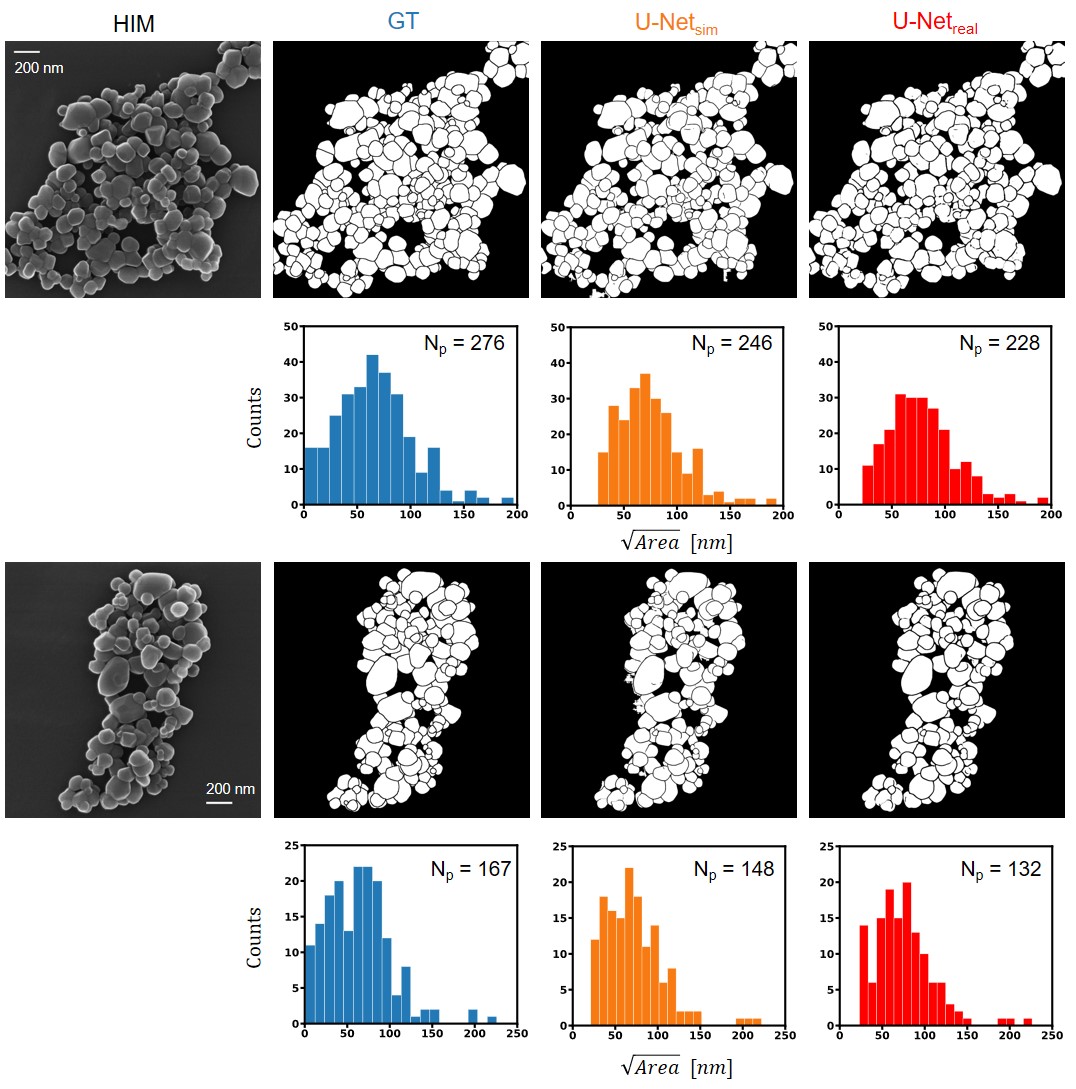}
\caption{\textbf{Post-processed segmentation results of U-Net\textsubscript{sim} and U-Net\textsubscript{real} for HIM SiO$_{2}$ nanoparticle images, compared to the ground truth (GT) annotation}. N\textsubscript{p} denotes the number of particles. Note that due to outlier correction in the post-processing, small objects are removed from the segmentation that fall below a certain size, which is also reflected in the histograms.}
\label{supp_fig:tio2_histograms_3}
\end{figure}

\clearpage
\begin{figure}[h!]
\centering
\includegraphics[scale=0.95]{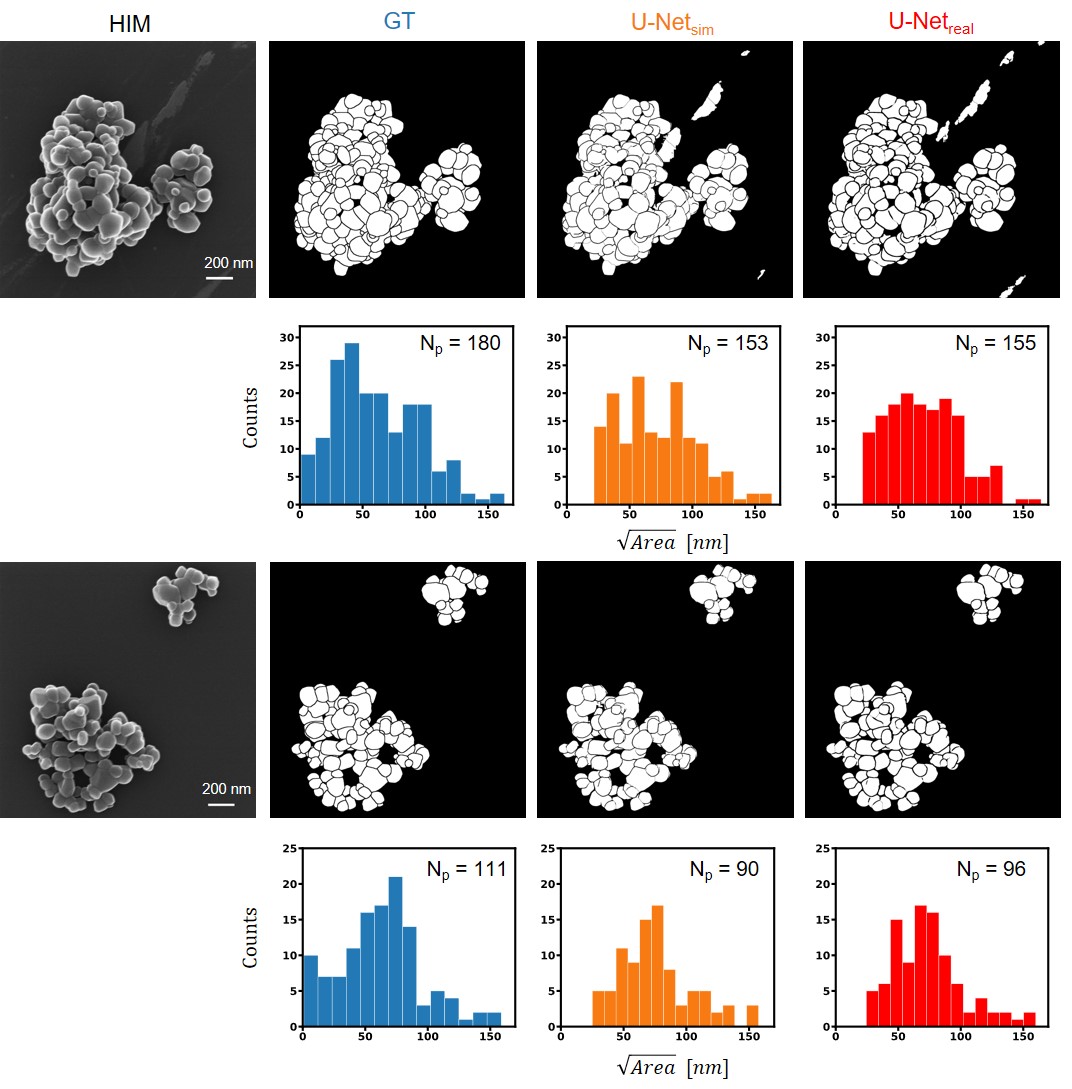}
\caption{\textbf{Post-processed segmentation results of U-Net\textsubscript{sim} and U-Net\textsubscript{real} for HIM SiO$_{2}$ nanoparticle images, compared to the ground truth (GT) annotation}. N\textsubscript{p} denotes the number of particles. Note that due to outlier correction in the post-processing, small objects are removed from the segmentation that fall below a certain size, which is also reflected in the histograms.}
\label{supp_fig:tio2_histograms_4}
\end{figure}

\clearpage
\section{Segmentation results of U-Net\texorpdfstring{\textsubscript{sim}}{sim}}
\subsection{SiO\texorpdfstring{\textsubscript{2}}{2} nanoparticles}
\begin{figure}[h!]
\centering
\includegraphics[scale=0.95]{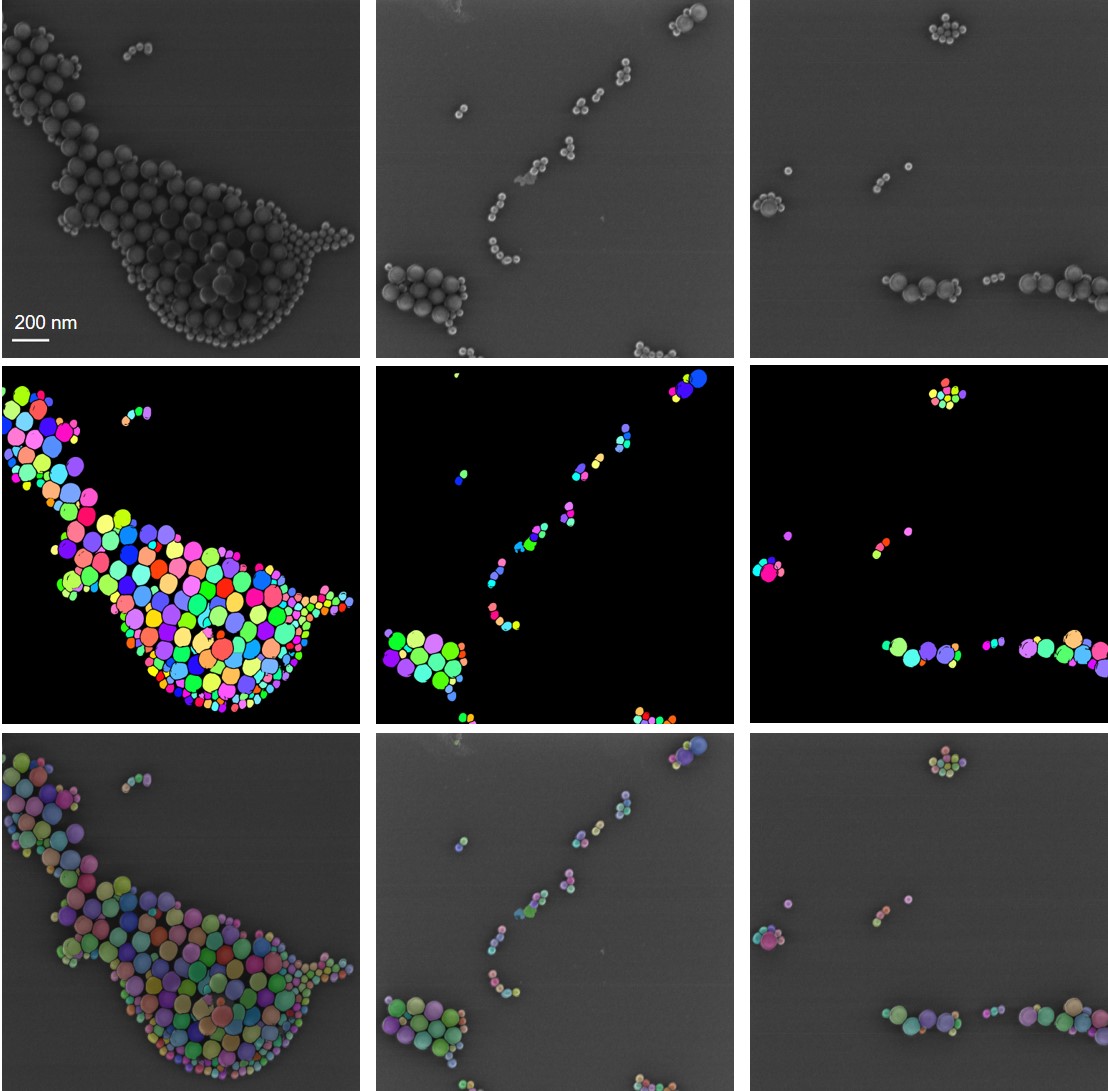}
\caption{\textbf{Segmentation results of U-Net\textsubscript{sim} for HIM SiO$_{2}$ nanoparticle images.} The top row displays real SiO$_{2}$ HIM images. The second row visualizes a connected component labeling (CCL) of the post-processed segmentation predicted by U-Net\textsubscript{sim}. Each color is associated with an individual particle. Note that due to the limited number of colors used for the CCL, neighboring particles may result in having the same color although not being connected. The bottom row shows an overlay of  HIM images with the CCL.}
\label{supp_fig:sio2_seg_1}
\end{figure}

\clearpage
\begin{figure}[h!]
\centering
\includegraphics[scale=0.95]{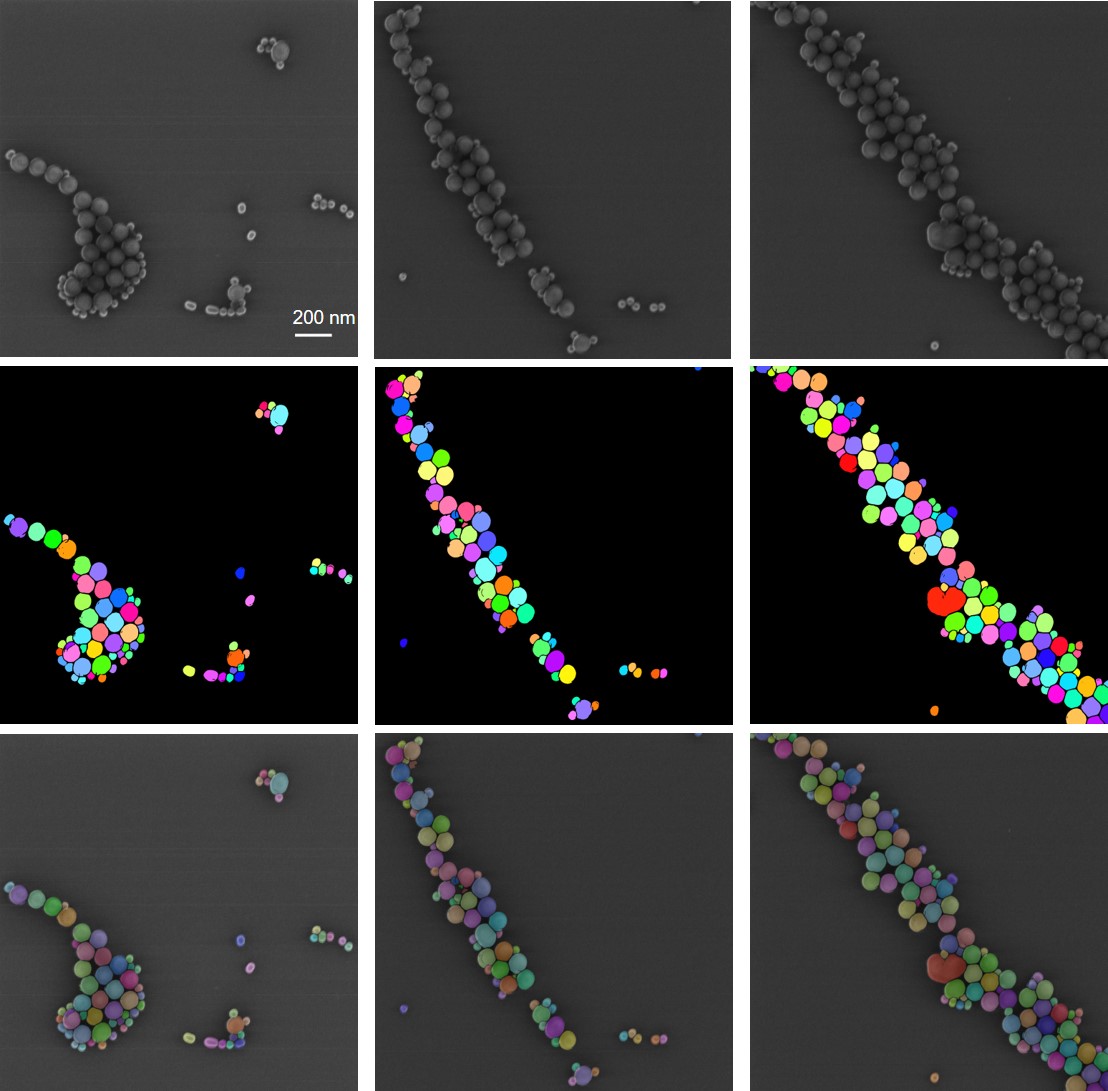}
\caption{\textbf{Segmentation results of U-Net\textsubscript{sim} for HIM SiO$_{2}$ nanoparticle images.} The top row displays real SiO$_{2}$ HIM images. The second row visualizes a connected component labeling (CCL) of the post-processed segmentation predicted by U-Net\textsubscript{sim}. Each color is associated with an individual particle. Note that due to the limited number of colors used for the CCL, neighboring particles may result in having the same color although not being connected. The bottom row shows an overlay of  HIM images with the CCL.}
\label{supp_fig:sio2_seg_2}
\end{figure}

\clearpage
\begin{figure}[h!]
\centering
\includegraphics[scale=0.95]{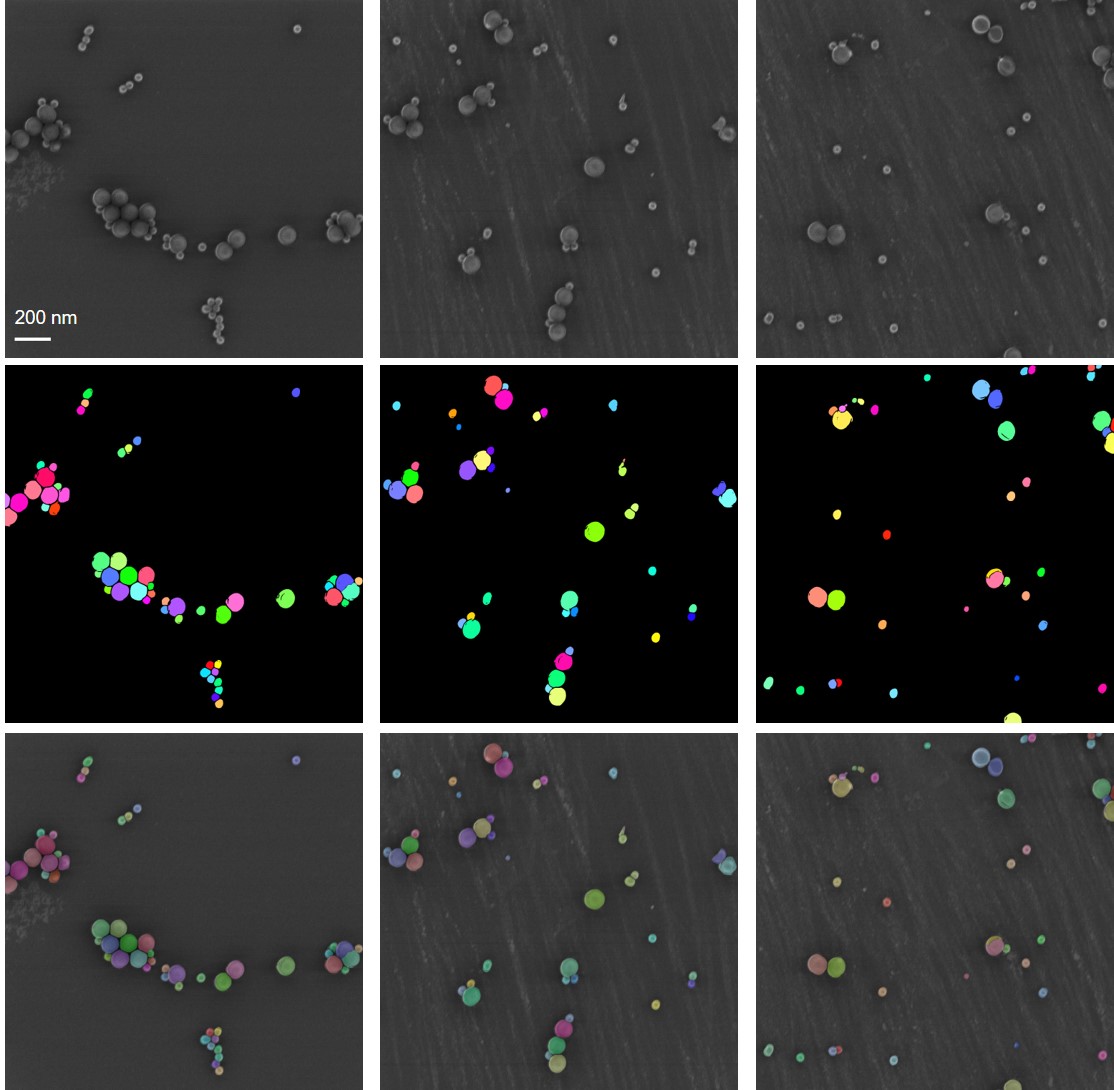}
\caption{\textbf{Segmentation results of U-Net\textsubscript{sim} for HIM SiO$_{2}$ nanoparticle images.} The top row displays real SiO$_{2}$ HIM images. The second row visualizes a connected component labeling (CCL) of the post-processed segmentation predicted by U-Net\textsubscript{sim}. Each color is associated with an individual particle. Note that due to the limited number of colors used for the CCL, neighboring particles may result in having the same color although not being connected. The bottom row shows an overlay of  HIM images with the CCL.}
\label{supp_fig:sio2_seg_3}
\end{figure}


\clearpage
\subsection{TiO\texorpdfstring{\textsubscript{2}}{2} nanoparticles}
\begin{figure}[h!]
\centering
\includegraphics[scale=0.95]{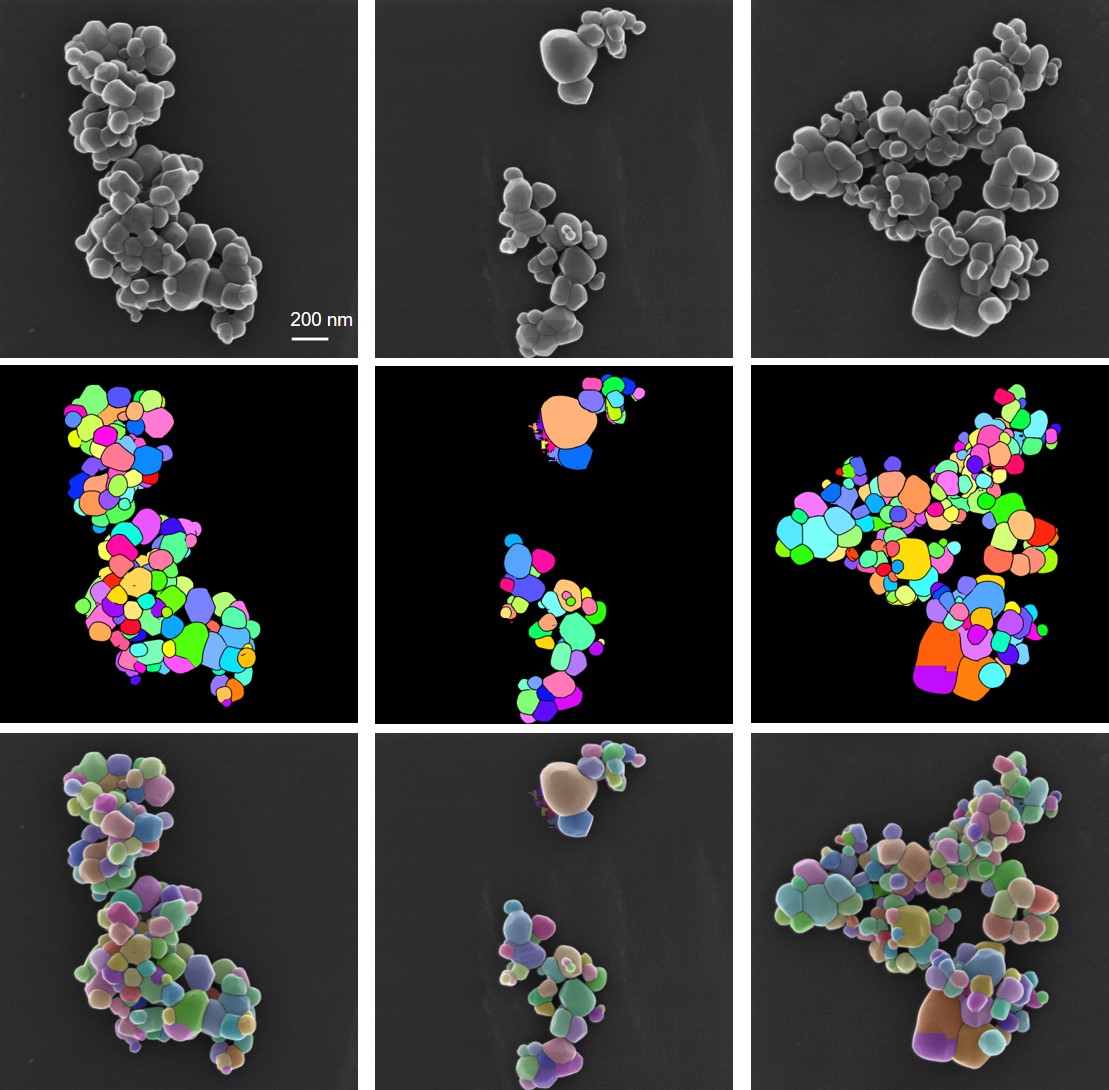}
\caption{\textbf{Segmentation results of U-Net\textsubscript{sim} for HIM TiO$_{2}$ nanoparticle images.} The top row displays real TiO$_{2}$ HIM images. The second row visualizes a connected component labeling (CCL) of the post-processed segmentation predicted by U-Net\textsubscript{sim}. Each color is associated with an individual particle. Note that due to the limited number of colors used for the CCL, neighboring particles may result in having the same color although not being connected. The bottom row shows an overlay of  HIM images with the CCL.}
\label{supp_fig:tio2_seg_1}
\end{figure}

\clearpage
\begin{figure}[h!]
\centering
\includegraphics[scale=0.95]{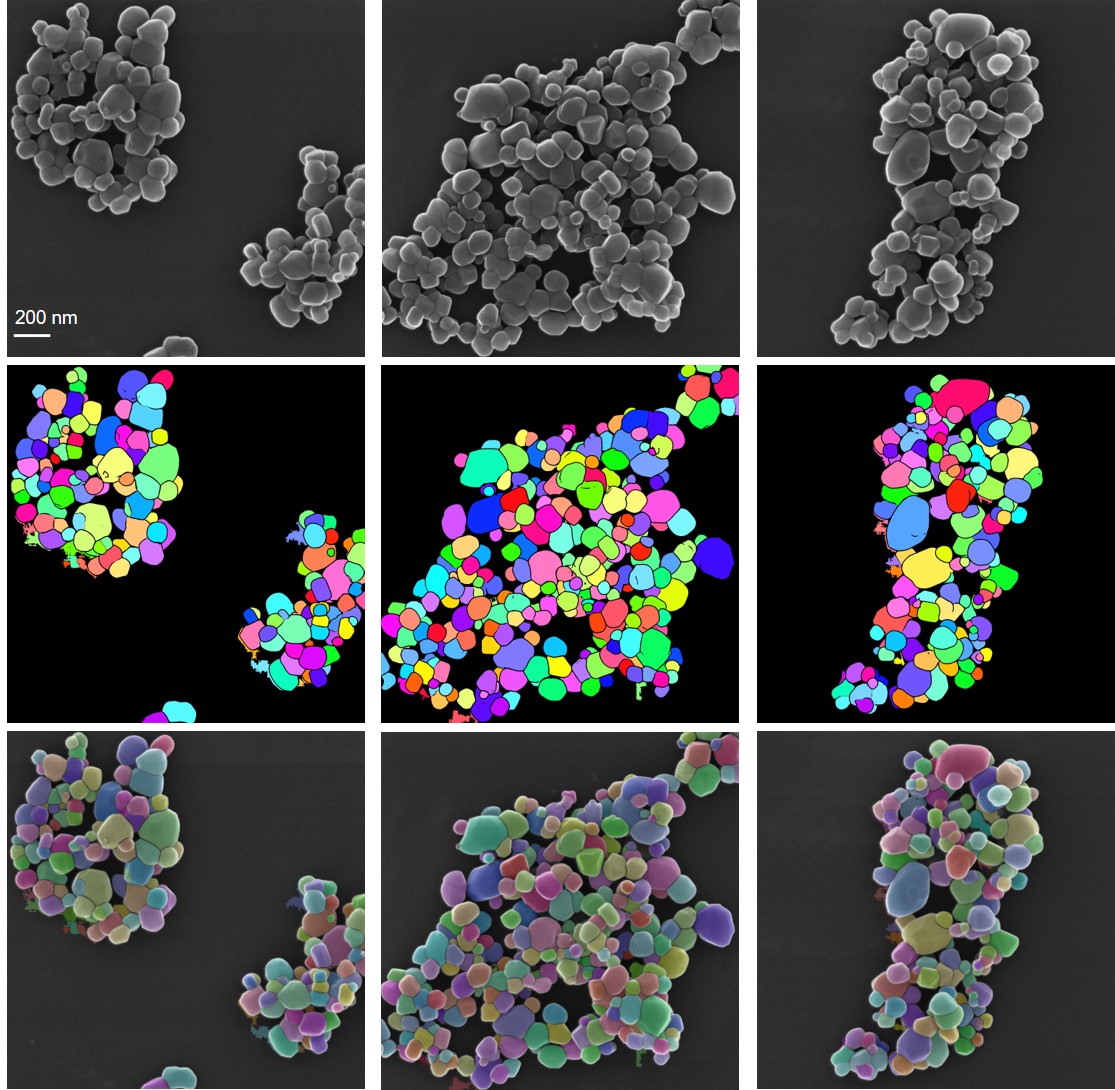}
\caption{\textbf{Segmentation results of U-Net\textsubscript{sim} for HIM TiO$_{2}$ nanoparticle images.} The top row displays real TiO$_{2}$ HIM images. The second row visualizes a connected component labeling (CCL) of the post-processed segmentation predicted by U-Net\textsubscript{sim}. Each color is associated with an individual particle. Note that due to the limited number of colors used for the CCL, neighboring particles may result in having the same color although not being connected. The bottom row shows an overlay of  HIM images with the CCL.}
\label{supp_fig:tio2_seg_2}
\end{figure}

\clearpage
\begin{figure}[h!]
\centering
\includegraphics[scale=0.95]{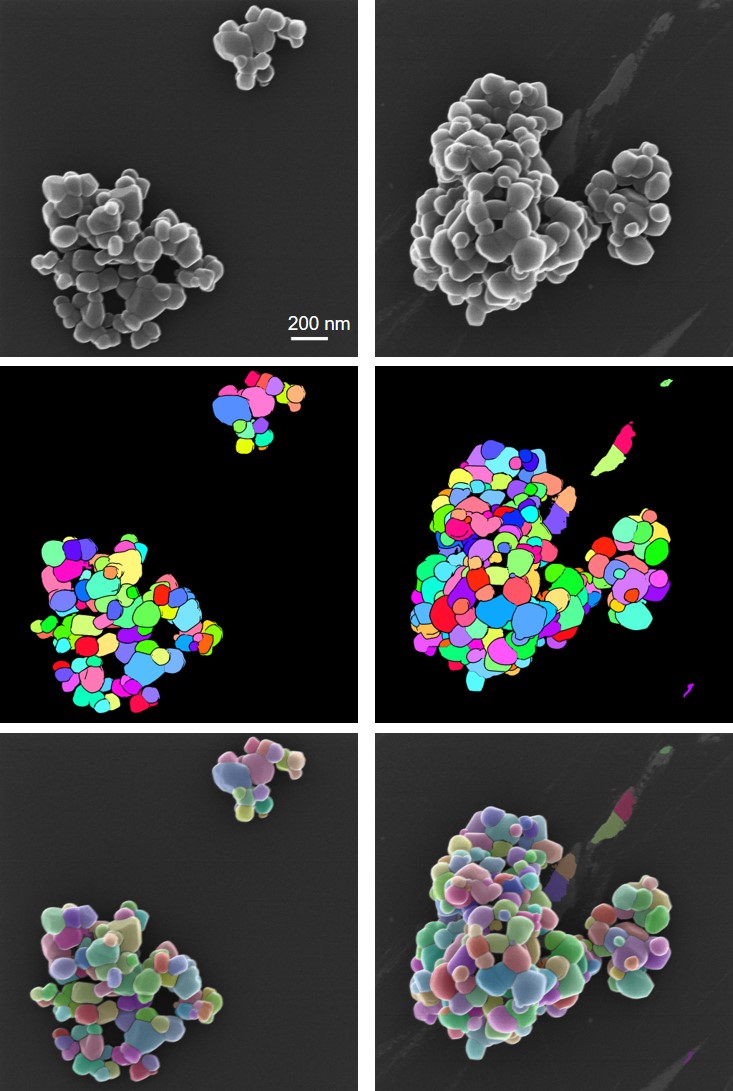}
\caption{\textbf{Segmentation results of U-Net\textsubscript{sim} for HIM TiO$_{2}$ nanoparticle images.} The top row displays real TiO$_{2}$ HIM images. The second row visualizes a connected component labeling (CCL) of the post-processed segmentation predicted by U-Net\textsubscript{sim}. Each color is associated with an individual particle. Note that due to the limited number of colors used for the CCL, neighboring particles may result in having the same color although not being connected. The bottom row shows an overlay of  HIM images with the CCL.}
\label{supp_fig:tio2_seg_3}
\end{figure}

\clearpage
\subsection{Ag}
\begin{figure}[h!]
\centering
\includegraphics[scale=0.95]{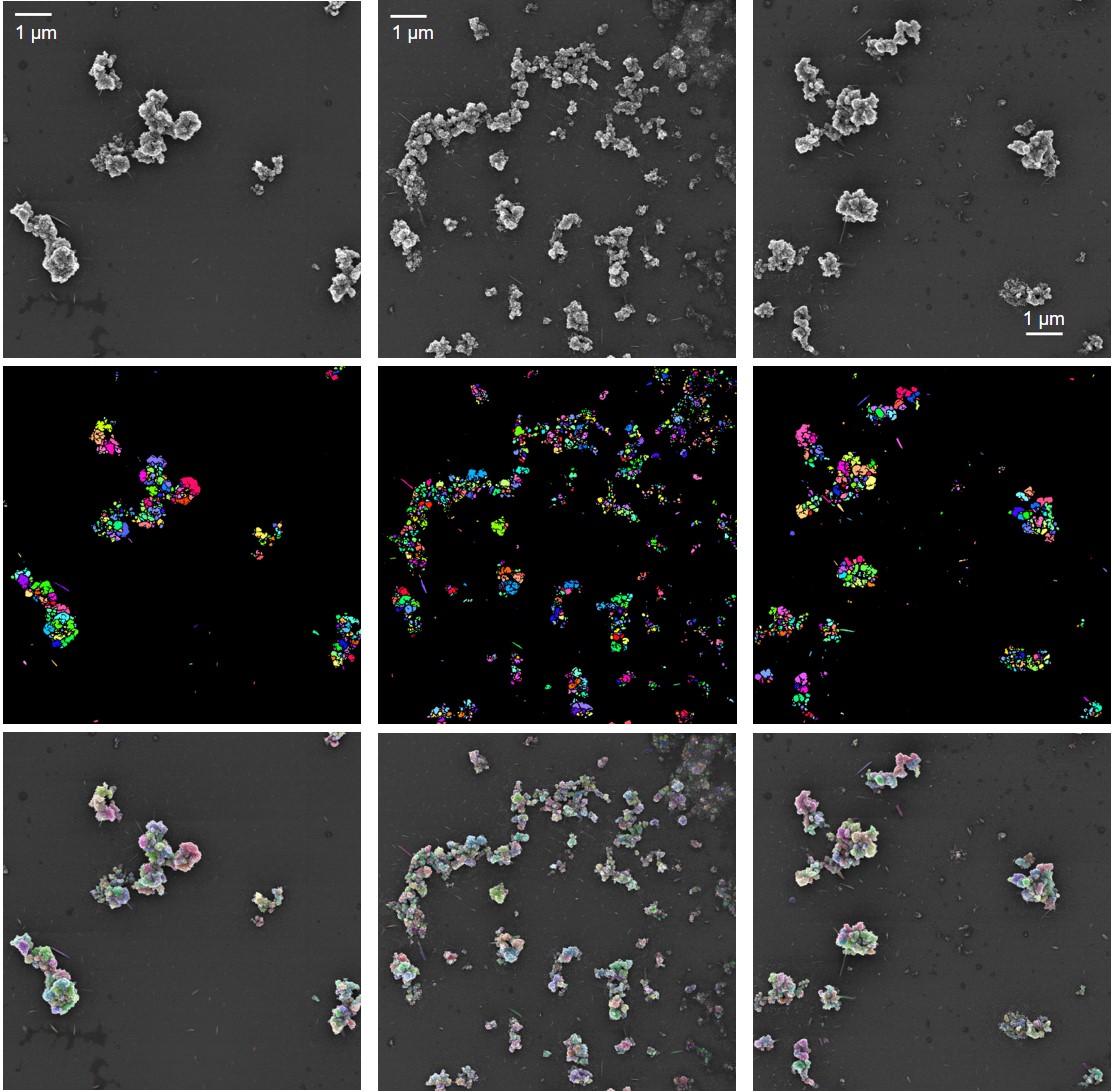}
\caption{\textbf{Segmentation results of U-Net\textsubscript{sim} for HIM Ag images.} The top row displays real Ag HIM images. The second row visualizes a connected component labeling (CCL) of the segmentation predicted by U-Net\textsubscript{sim} without post-processing. Each color is associated with an individual particle. Note that due to the limited number of colors used for the CCL, neighboring objects may result in having the same color although not being connected. The bottom row shows an overlay of  HIM images with the CCL.}
\label{supp_fig:ag_seg_1}
\end{figure}

\clearpage
\begin{figure}[h!]
\centering
\includegraphics[scale=0.95]{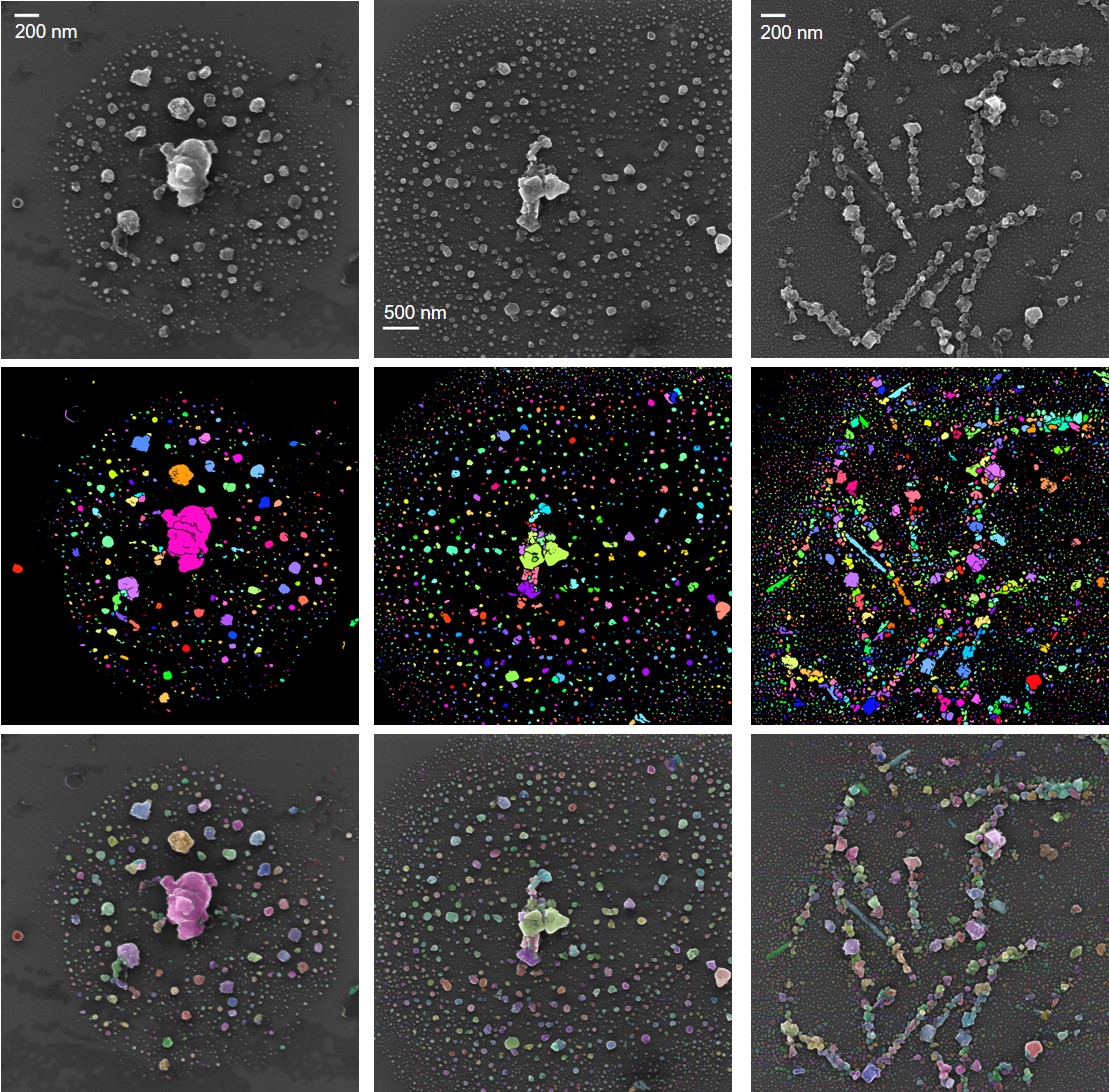}
\caption{\textbf{Segmentation results of U-Net\textsubscript{sim} for HIM Ag images.} The top row displays real Ag HIM images. The second row visualizes a connected component labeling (CCL) of the segmentation predicted by U-Net\textsubscript{sim} without post-processing. Each color is associated with an individual particle. Note that due to the limited number of colors used for the CCL, neighboring objects may result in having the same color although not being connected. The bottom row shows an overlay of  HIM images with the CCL.}
\label{supp_fig:ag_seg_2}
\end{figure}

\clearpage
\begin{figure}[h!]
\centering
\includegraphics[scale=0.95]{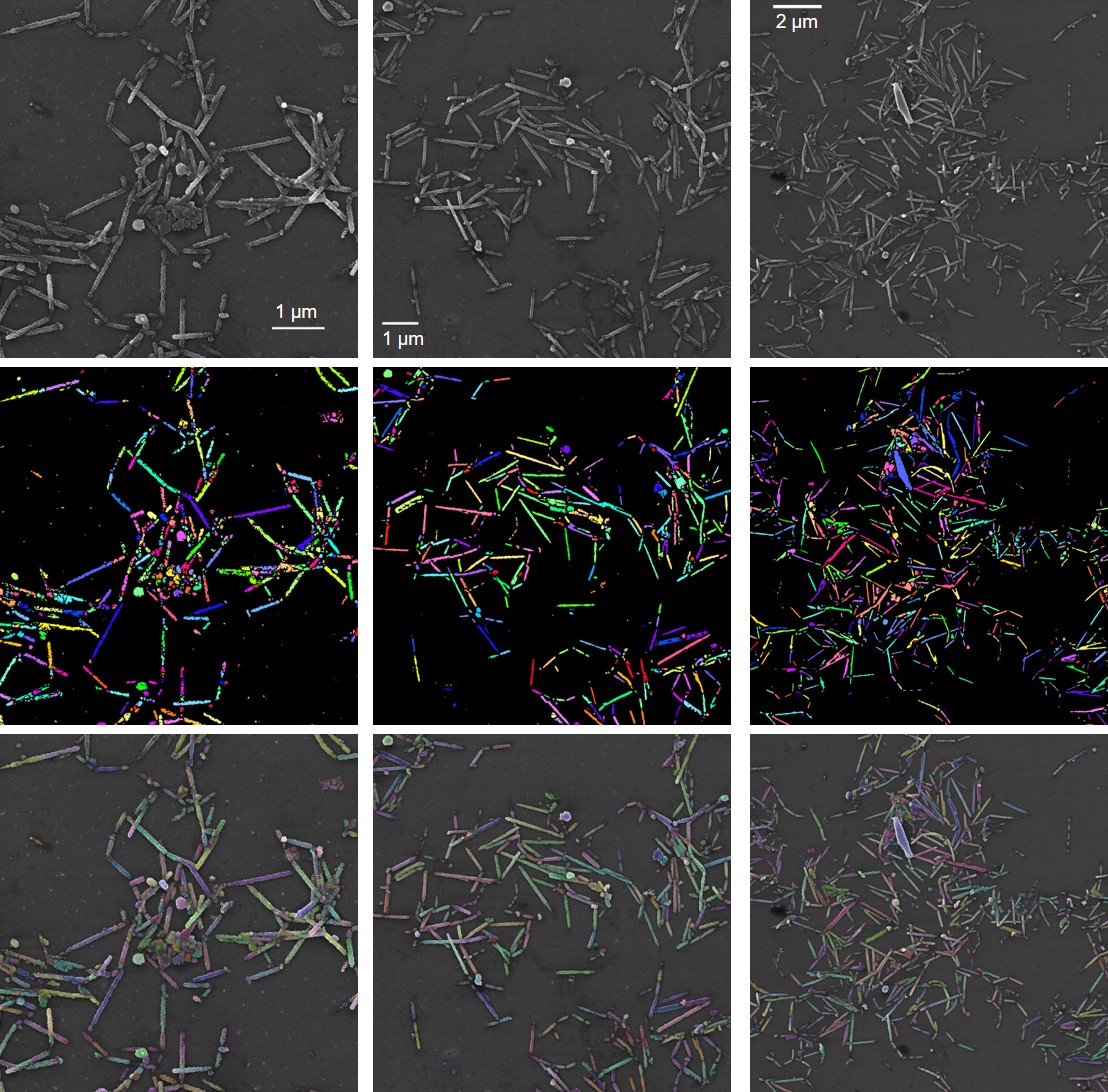}
\caption{\textbf{Segmentation results of U-Net\textsubscript{sim} for HIM Ag images.} The top row displays real Ag HIM images. The second row visualizes a connected component labeling (CCL) of the segmentation predicted by U-Net\textsubscript{sim} without post-processing. Each color is associated with an individual particle. Note that due to the limited number of colors used for the CCL, neighboring objects may result in having the same color although not being connected. The bottom row shows an overlay of  HIM images with the CCL.}
\label{supp_fig:ag_seg_3}
\end{figure}

\end{document}